\documentclass[10pt,journal,compsoc]{IEEEtran}

\ifCLASSOPTIONcompsoc
  \usepackage[nocompress]{cite}
\else
  \usepackage{cite}
\fi

\usepackage{times}
\usepackage{helvet}
\usepackage{graphicx}
\usepackage{subfigure}
\usepackage{amsmath}
\usepackage{amssymb}
\usepackage{theorem}
\usepackage{epstopdf}
\usepackage{color}

\newtheorem{lemma}{Lemma}

\newtheorem{theorem}{Theorem}
\newtheorem{proposition}{Proposition}
\theorembodyfont{\upshape}
\usepackage{algorithm}
\usepackage{algorithmic}
\usepackage{float}
\usepackage{courier}
\newcommand{\EE}{\mathbb{E}}
\newcommand{\RR}{\mathbb{R}}

\ifCLASSINFOpdf
\else
\fi

\begin{document}

\title{Training Deep Neural Networks with Adaptive Momentum Inspired by the Quadratic Optimization}

\author{Tao~Sun, Huaming Ling, Zuoqiang Shi, Dongsheng Li, and Bao Wang
\IEEEcompsocitemizethanks{
\IEEEcompsocthanksitem T. Sun and Dongsheng Li are  with College of Computer, National University of Defense Technology (nudtsuntao@163.com;dsli@nudt.edu.cn).
\IEEEcompsocthanksitem H. Ling and Z. Shi are with Yau Mathematical Sciences Center, Tsinghua University (linghm18@mails.tsinghua.edu.cn;zqshi@mail.tsinghua.edu.cn).
\IEEEcompsocthanksitem B. Wang is with Department of Mathematics, University of Utah (wangbaonj@gmail.com).
\IEEEcompsocthanksitem B. Wang, D. Li and Z. Shi are the co-corresponding authors.
\IEEEcompsocthanksitem The first and second authors contributed equally.}
\thanks{This work  is sponsored in part by National Key R\&D Program of China (2018YFB0204300), and the National Science Foundation of China (No. 61906200).}}


\IEEEtitleabstractindextext{%
\begin{abstract}
Heavy ball momentum is crucial in accelerating (stochastic) gradient-based optimization algorithms for machine learning. Existing heavy ball momentum is usually weighted by a uniform hyperparameter, which relies on excessive tuning. Moreover, the calibrated fixed hyperparameter may not lead to optimal performance. In this paper, to eliminate the effort for tuning the momentum-related hyperparameter, we propose a new adaptive momentum inspired by the optimal choice of the heavy ball momentum for quadratic optimization. Our proposed adaptive heavy ball momentum can improve stochastic gradient descent (SGD) and Adam. SGD and Adam with the newly designed adaptive momentum are more robust to large learning rates, converge faster, and generalize better than the baselines. We verify the efficiency of SGD and Adam with the new adaptive momentum on extensive machine learning benchmarks, including image classification, language modeling, and machine translation. Finally, we provide convergence guarantees for SGD and Adam with the proposed adaptive momentum.
\end{abstract}

}

\maketitle

\IEEEdisplaynontitleabstractindextext

%
\IEEEpeerreviewmaketitle

 \section{Introduction}
\label{sec:intro}
\IEEEPARstart{C}{onsider}  the following empirical risk minimization (ERM) problem
\begin{equation}\label{eq:finite-sum}
\min_{{\bf x}\in \RR^d}\{f({\bf x}):=\frac{1}{n}\sum_{i=1}^n f_i({\bf x})=\frac{1}{n}\sum_{i=1}^n \mathcal{L}(\sigma({\bf s}_i;{\bf x}),y_i)\},
\end{equation}
where $\mathcal{L}$ is the loss function, $\sigma$ is the machine learning model parameterized by ${\bf x}$, and $({\bf s}_i,y_i)$ is a 
sample-label pair.
Stochastic gradient descent (SGD) \cite{robbins1951stochastic} is a simple yet effective algorithm to solve \eqref{eq:finite-sum}, which updates ${\bf x}$ according to
$$
{\bf x}^{k+1}={\bf x}^k-\gamma {\bf g}^k,
$$
where $\gamma>0$ is the learning rate and ${\bf g}^k=\frac{1}{m}\sum_{i=1}^m\nabla f_i({\bf x}^k)$ with $m\ll n$ is the 
mini-batch stochastic gradient of $f$.

Heavy ball (HB) algorithm \cite{polyak1964some} leverages memory to accelerate GD, which updates ${\bf x}$ as follows
\begin{equation}\label{eq:HB-intro}
{\bf x}^{k+1} = {\bf x}^k-\gamma {\bf g}^k+\beta({\bf x}^k-{\bf x}^{k-1}),
\end{equation}
where $0\leq \beta< 1$ is the momentum hyperparameter. 
By introducing  momentum states $({\bf m}^k)_{k\geq 0}$, we can rewrite HB as
$$
{\bf m}^{k+1} = \beta{\bf m}^k+{\bf g}^k;\ {\bf x}^{k+1}={\bf x}^k-\gamma{\bf m}^{k+1}.
$$
If we integrate the adaptive learning rate with the heavy ball algorithm and rescale ${\bf m}^k$ and ${\bf g}^k$, we obtain the following celebrated Adam algorithm \cite{kingma2014adam} \footnote{For the sake of simplicity, we omit the bias correction here.}
\begin{equation}\label{eq:Adam-intro}
      \begin{aligned}
    {\bf x}^{k+1}& = {\bf x}^k-\gamma {{\bf m}^{k}}/{[{\bf v}^{k}]^{1/2}};\\
{\bf m}^{k+1} &= \beta{\bf m}^k + (1-\beta){\bf g}^k; \\
{\bf v}^{k+1} &= \alpha{\bf v}^k + (1-\alpha)[{\bf g}^k]^2,\\
\end{aligned}
\end{equation}
where $\alpha,\beta,\gamma >0$ are three positive constants, and $[\cdot]^2$ and $[\cdot]^{1/2}$ denote the element-wise square and squared-root, respectively. These first-order 
algorithms are among the methods of choice for signal processing \cite{beck2009fast} and machine learning; in particular, for training deep neural networks (DNNs) \cite{bottou2018optimization}.

The momentum in both HB and Adam is weighted by a constant $\beta$. A fine-tuned $\beta$ is crucial for accelerating SGD and Adam in 
training DNNs and improving their generalization \cite{krizhevsky2009learning,sutskever2013importance,keskar2017improving,wilson2017marginal,luo2019adaptive}.
Indeed, \cite{polyak1964some,ghadimi2015global,wang2020provable} establish the theoretical acceleration of momentum for convex optimization and training one-layer neural networks, and these results depend on   specialized 
choices of $\beta$ and $\gamma$.
Calibrating the momentum hyperparameter $\beta$ is computationally expensive. Moreover, training machine learning models may require different 
$\beta$ in different iterations.
To the best of our knowledge, there is no principled way to choose the optimal $\beta$ for training DNNs. Therefore, it is natural and important to ask the following question:

\emph{Can we design an adaptive momentum without calibrating for the momentum-related hyperparameter to improve GD/SGD and Adam with 
convergence guarantees? }

\vspace{-0.3cm}
\subsection{Contributions}
We answer the above question affirmatively by replacing $\beta$ in HB \eqref{eq:HB-intro} and Adam \eqref{eq:Adam-intro} with the following iteration-dependent adaptive scheme
\begin{equation}\label{eq:adp:momentum}
{\footnotesize \beta_{k+1} = \begin{cases}
{\bf Proj}_{[0,1-\delta]}\left(\left[1-\sqrt{\gamma\frac{{\|\bf g}^k-{\bf g}^{k-1}\|}{\|{\bf x}^k-{\bf x}^{k-1}\|}} \right]^2 \right),&k\geq 2,\\
0,&k=0,1,
\end{cases}
}
\end{equation}
where $\|\cdot\|$ denotes the $\ell_2$ norm of a given vector,
and ${\bf Proj}_{[0,1-\delta]}(\cdot):=\max(0,\min(\cdot,1-\delta))$ with $\delta$ be the threshold, \emph{which is simply set to $10^{-3}$ in all the following experiments.} We will study the effects of $\delta$ in 
Appendix~\ref{sec:Ablation}.

We start from a fine-grained convergence analysis of HB for quadratic optimization, resulting in the optimal choice for momentum. Then, we design a simple sequence to approximate the optimal momentum
and result in our proposed adaptive momentum \eqref{eq:adp:momentum}.
The convergence of HB, proximal HB, and Adam with adaptive momentum in both convex and nonconvex settings is easy to guarantee; as a minor contribution, we show the convergence of the adaptive momentum schemes in various settings.
Finally, we perform extensive experiments to show that these adaptive momentum schemes are more robust to larger learning rates, converge faster, and generalize better than the baselines.

\subsection{Additional Related Works}
The advances of momentum-based acceleration have been booming since the pioneering work of \cite{polyak1964some,nesterov1983method}. In this section, we briefly review some of the most exciting and related works on developing and analyzing HB and Nesterov's acceleration (a class of adaptive momentum). Also, we will discuss some of the recent development of adaptive momentum. 

\paragraph{Heavy ball} The convergence of deterministic HB, i.e., HB with exact gradient, has been thoroughly studied by \cite{ochs2014ipiano,ghadimi2015global,ochs2015ipiasco,sun2019non,sun2020nonergodic} in both convex and nonconvex cases. An interesting finding is that HB can escape saddle points in nonconvex optimization by using a larger learning rate than GD \cite{sun2019heavy}. HB momentum has also been successfully integrated into SGD to improve training DNNs. Especially for image classification  \cite{krizhevsky2009learning,sutskever2013importance,bottou2018optimization}. The authors of \cite{yan2018unified} propose and analyze a generalized stochastic momentum scheme. In \cite{ma2018quasi,gitman2019understanding}, the authors develop quasi-hyperbolic momentum to accelerate SGD; the quasi-hyperbolic momentum averages a plain SGD step with a momentum step. The effects of momentum have also been studied in large batch size scenario \cite{NEURIPS2019_e0eacd98}.
SGD with HB momentum shows good empirical performance in both accelerating convergence and improving generalization, but the theoretical acceleration guarantee is still unclear.

\paragraph{Nesterov's acceleration} Nesterov's acceleration can be obtained by
replacing $\beta$ with a specially designed time-varying formula, which is a special adaptive momentum.
Nesterov's acceleration achieves the optimal convergence rate for smooth convex minimization with access to gradient information only \cite{nesterov1983method,nesterov2013introductory}. The first use of Nesterov's acceleration in stochastic gradient is proposed by \cite{wiegerinck1994stochastic} but without any theoretical guarantee.  
The convergence of Nesterov's acceleration can be proved for 
stochastic gradient using diminishing learning rates \cite{JMLR:v17:16-157,yan2018unified}. In \cite{kulunchakov2019generic,aybat2020robust}, the authors prove that the stochastic Nesterov's acceleration converges to a neighborhood of the minimum. Nesterov's acceleration for over-parameterized models with a diminishing momentum has been studied in \cite{vaswani2019fast}.
Nevertheless, in the finite-sum setting, the Nesterov's acceleration is proved to possibly diverge with the usual choice of the learning rate and momentum \cite{liu2019accelerating,assran2020convergence}. The possible divergence of stochastic Nesterov's acceleration limits its use for deep learning. To this end, a restart momentum is employed for the stochastic Nesterov's acceleration \cite{wang2020scheduled}, which achieves further acceleration from the numerical tests, but the convergence is proved with non-trivial assumptions.

\paragraph{Other variants of HB and Nesterov's acceleration} Many other momentum schemes have been developed to fix the possible divergence issue of the Nesterov's acceleration \cite{lan2012optimal,ghadimi2012optimal,ghadimi2013optimal,allen2017katyusha,cohen2018acceleration,liu2019accelerating}.
However, these algorithms are not popular in deep learning.

\paragraph{Other adaptive momentum} The authors in \cite{wang2020stochastic} develop a class of nonlinear conjugate gradient (NCG)-style adaptive momentum for HB,
and they prove acceleration for quadratic function in the deterministic case and numerically show the effectiveness 
in training DNNs.

%
%

\subsection{Notation}
We denote scalars/vectors by lower case/lower case boldface letters.
We denote matrices by upper case boldface letters. For 
${\bf x} = (x_1, \cdots, x_d)\in \mathbb{R}^d$,
  we use $\|{\bf x}\|$ to denote its $\ell_2$ norm. 
  For a matrix ${\bf A}$, we use ${\bf A}^\top$ to denote its transpose and use $\|{\bf A}\|_p$ to denote its induced norm by the vector $\ell_p$ norm. Given two sequences $\{a_n\}$ and $\{b_n\}$, we write $a_n=\mathcal{O}(b_n)$ if there exists
a positive constant $0<C<+\infty$ such that
$a_n \leq C b_n$, and we write $a_n=\Theta(b_n)$ if there exist two positive constants $C_1$ and $C_2$ such that $a_n \leq C_1 b_n$ and $b_n \leq C_2 a_n$.
We denote the set $\{1, 2, \cdots, m\}$ as $[m]$. We denote the gradient and Hessian of a function $f({\bf x}): \mathbb{R}^d \rightarrow \mathbb{R}$ by $\nabla f({\bf x})$ and $\nabla^2 f({\bf x})$, respectively.

\section{Motivation and Algorithms}\label{sec:algorithms}

To get a better understanding of HB\footnote{Here, HB means GD with HB momentum.}, we start from the quadratic problem.
In particular, we have
\begin{lemma}\label{th1}
Let $f$ be a quadratic function with $\nabla f({\bf x})= {\bf A}{\bf x}+{\bf b}$, where ${\bf A}$ is positive definite. Moreover, we denote the smallest ($\lambda_{\min}({\bf A})$) and the largest eigenvalues ($\lambda_{\max}({\bf A})$) of ${\bf A}$ as $\nu$ and $L$, respectively. Then, given any fixed $\gamma\leq{1}/{L}$, the optimal choice for $\beta$ is $\beta=(1-\sqrt{\gamma\nu})^2$. In this case, HB achieves a convergence rate of
$\|{\bf x}^{k}-{\bf x}^*\|\leq (1-\sqrt{\gamma\nu}+\epsilon)^k\cdot 2\|{\bf x}^{0}-{\bf x}^*\|$
as $k\geq K({\bf A},\epsilon)$, where $\epsilon>0$ is very small and $K({\bf A},\epsilon)\in\mathbb{Z}^+$ is an integer depends on ${\bf A}$ and $\epsilon$.
\end{lemma}

Compared with \cite{recht2010cs726}, Lemma~\ref{th1} is an improved result 
for HB methods. In particular, Lemma~\ref{th1} shows that the optimal hyperparameter for the HB momentum should be $(1-\sqrt{\gamma\nu})^2$ if $\gamma\leq 1/L$. However, the smallest eigenvalue $\nu$ is unknown. Therefore, we construct the sequence,
$\|{\bf g}^k-{\bf g}^{k-1}\|/\|{\bf x}^k-{\bf x}^{k-1}\|$, to approximate $\nu$. It is easy to check that $\|{\bf g}^k-{\bf g}^{k-1}\|/\|{\bf x}^k-{\bf x}^{k-1}\|\in [\nu,L]$. Moreover, we have the following lemma, which shows that $\|{\bf g}^k-{\bf g}^{k-1}\|/\|{\bf x}^k-{\bf x}^{k-1}\|\rightarrow \nu$.

\begin{lemma}\label{le1}
Assume that the conditions in Lemma \ref{th1} hold and $\{{\bf x}^k\}_{k\geq 0}$ is generated by HB \eqref{eq:HB-intro}. Also, assume ${\bf A}$ has a unique minimum singular value. If $\gamma\leq {1}/{L}$, for any fixed $0\leq \beta<1$, we have
$$ \lim_{k\rightarrow \infty}\frac{\|{\bf g}^k-{\bf g}^{k-1}\|}{\|{\bf x}^k-{\bf x}^{k-1}\|}=\nu.$$
\end{lemma}
\begin{figure}
\vspace{-0.15 in}
\includegraphics[width=0.9\linewidth]{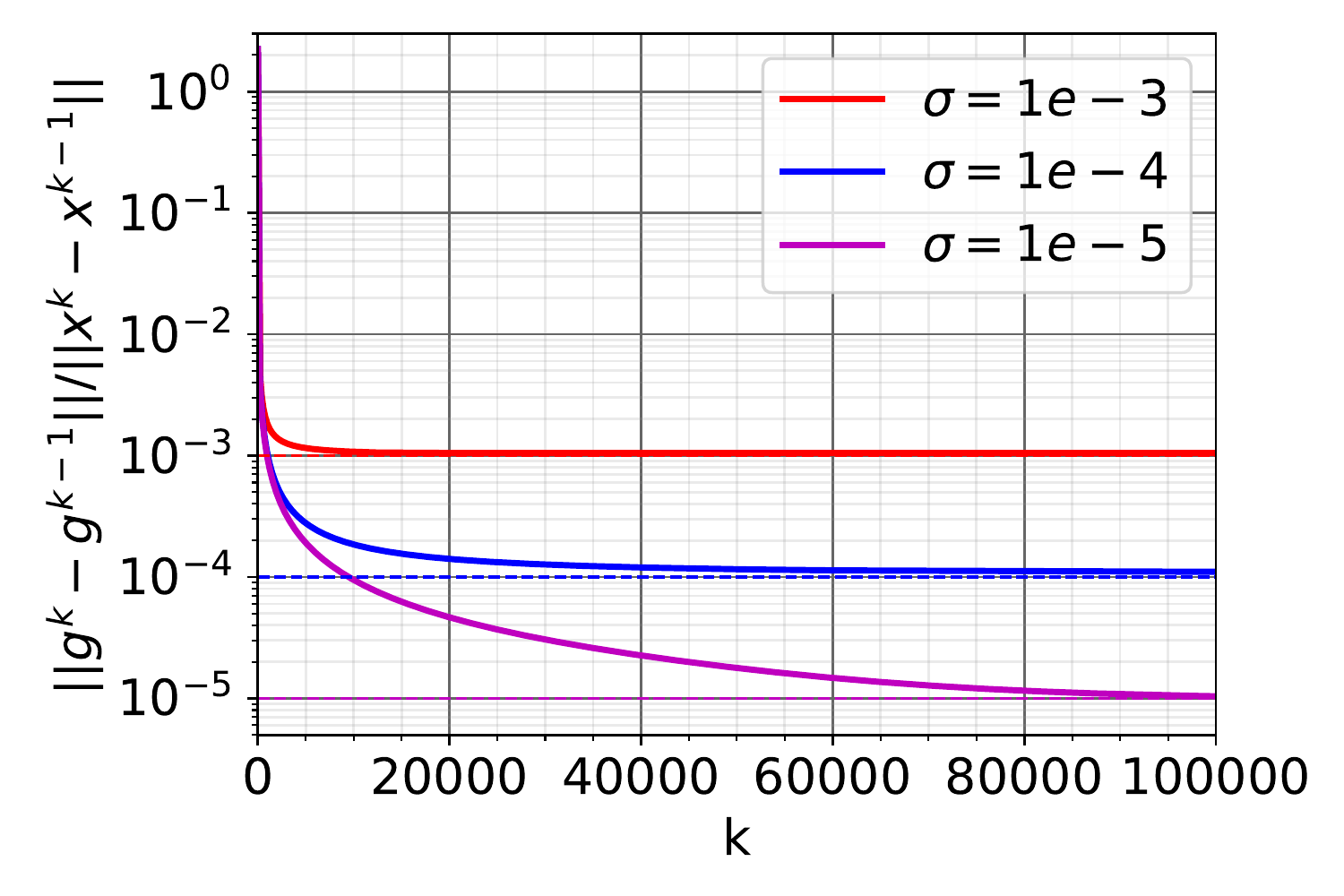}
\vspace{-0.3in}
\caption{ Iteration ($k$) vs. $\|{\bf g}^{k}-{\bf g}^{k-1}\|/\|{\bf x}^{k}-{\bf x}^{k-1}\|$ when HB \eqref{eq:HB-intro} is applied to solve \eqref{eq:quadratic:Laplacian} with different $\sigma$. It is evident that $\|{\bf g}^{k}-{\bf g}^{k-1}\|/\|{\bf x}^{k}-{\bf x}^{k-1}\|\rightarrow \sigma$.}\label{fig:verification:lemma1}
\end{figure}
To numerically verify Lemma~\ref{le1}, we apply HB \eqref{eq:HB-intro} to solve the following quadratic 
problem
\begin{equation}\label{eq:quadratic:Laplacian}
\min_{\bf x\in \mathbb{R}^d} {\bf x}^T(\sigma {\bf I} +{\bf L}){\bf x} - {\bf b}^T{\bf x},
\end{equation}
where ${\bf I}$ is the $d\times d$ identity matrix, ${\bf L}$ is the Laplacian matrix of a cyclic graph (see Appendix~\ref{appendix-laplacian} for details),
${\bf b} \in \mathbb{R}^d$ is a vector whose first entry is $1$ and all the other entries are $0$s.

It is evident that $\nu:=\lambda_{\min}(\sigma {\bf I}+{\bf L})=\sigma$. We consider three different $\sigma$: $10^{-3}, 10^{-4}$ and $10^{-5}$.\footnote{We select these $\sigma$s such that the condition number of $\sigma{\bf I}+{\bf L}$ is very large, slowing convergence of the gradient-based algorithm,
which is helpful to visualize the convergence behavior of $\|{\bf g}^k-{\bf g}^{k-1}\|/\|{\bf x}^k-{\bf x}^{k-1}\|$.} Figure~\ref{fig:verification:lemma1} shows that $\|{\bf g}^k-{\bf g}^{k-1}\|/\|{\bf x}^k-{\bf x}^{k-1}\|$ converges to $\nu$ for all $\sigma$s above, where we set $\gamma=0.1$ and $\beta=0.9$.

\paragraph{HB with Adaptive Momentum} The previous result indicates that $([1-\sqrt{\gamma{\|{\bf g}^k-{\bf g}^{k-1}\|}/{\|{\bf x}^k-{\bf x}^{k-1}\|}}]^2)$ is a reasonable approximation to the optimal momentum for HB for quadratic optimization when $\gamma\leq 1/L$. We extend the above approximated optimal momentum to general objectives and stochastic HB, which gives us
\begin{subequations}\label{scheme}
\small
\hskip -0.2cm\begin{align}
{\bf x}^{k+1}&={\bf x}^k-\gamma{\bf g}^k+\beta_k({\bf x}^k-{\bf x}^{k-1}),\\
\beta_{k+1} &={\bf Proj}_{[0,1-\delta]}\left(\Big[1-\sqrt{\gamma\frac{\|{\bf g}^k-{\bf g}^{k-1}\|}{\|{\bf x}^k-{\bf x}^{k-1}\|}}\Big]^2\right),\label{schemeAup}
\end{align}
\end{subequations}
where ${\bf g}^k$ is $\nabla f({\bf x}^k)$ or an unbiased estimate of $\nabla f({\bf x}^k)$. In \eqref{schemeAup}, we project $\beta_{k+1}$ into $[0,1-\delta]$ to force the estimated momentum in a valid range. We name the above algorithm adaptive stochastic heavy ball (ASHB) when the stochastic gradient is used; we summarize ASHB in Algorithm~\ref{alg1}.
The update of $\beta_{k+1}$ merely involves a few additional algebraic computations. Thus, compared with SGD with constant momentum, \eqref{scheme} does not add much computational overhead.

\begin{algorithm}[H]
\caption{Adaptive Stochastic Heavy Ball (ASHB)}
\begin{algorithmic}\label{alg1}
\REQUIRE   parameters $\gamma>0,\delta>0$, integer $m\in \mathbb{Z}^+$\\
\textbf{Initialization}: ${\bf g}^0=\textbf{0}$,  ${\bf x}^0={\bf x}^1=\textbf{0}$, $\beta_0=0$\\
\textbf{for}~$k=1,2,\ldots$ \\
~~~ \textbf{step 1}: get ${\bf g}^k=\sum_{j=1}^m \nabla f_{i_j}({\bf x}^k)/m$\\
~~~ \textbf{step 2}: update as \eqref{scheme} \\
\textbf{end for}\\
\end{algorithmic}
\end{algorithm}

\section{Extend to Proximal Algorithms and Adam }\label{sec:Generalization}
In this section, we present two extensions of ASHB.
One is the proximal algorithm for 
composite optimization, and the other one is Adam for deep learning.
\subsection{Extend to Proximal Algorithms}
Proximal algorithms are used for solving the following composite optimization
\begin{equation}\label{composite}
\min\{ F({\bf x}):= f({\bf x})+g({\bf x}) \},\ {\bf x}\in \RR^d,
\end{equation}
where we assume that the proximal map of the function $\gamma g, \gamma>0$ is easy to obtain. In particular, when
$
\textbf{Prox}_{\gamma g}(\cdot):=\textrm{arg}\min_{{\bf x}}\{\gamma g({\bf x})+\frac{\|{\bf x}-\cdot\|^2}{2}\}
$
has a closed form solution. Composite optimization has been widely used in image processing
and statistics \cite{daubechies2004iterative,tibshirani1996regression,hale2007fixed}. To extend adaptive momentum to proximal gradient descent, we leverage the following approximation of the optimal momentum as  \eqref{schemeAup}
with ${\bf g}^k=\nabla f({\bf x}^k)$ for $f({\bf x})$ in \eqref{composite}. We call the resulting algorithm proximal adaptive heavy ball (PAHB), which is summarized in Algorithm~\ref{alg2}.

\subsection{Extend to Adam Algorithms}
Adam \cite{kingma2014adam} can be understood as Ada (adaptive learning rate) with momentum. We can further integrate adaptive momentum \eqref{eq:adp:momentum} with Adam, and we call the resulting algorithm Ada$^2$m. Again, 
we omit the bias correction steps for the sake of presentation. We summarize Ada$^2$m in Algorithm \ref{alg4}.
  \vspace{0pt}
\begin{algorithm}[H]

\caption{Proximal Adaptive Heavy Ball}
\begin{algorithmic}\label{alg2}
\REQUIRE   parameters $\gamma>0$, $\delta>0$\\
\textbf{Initialization}:   ${\bf x}^0={\bf x}^1=\textbf{0}$, $\beta_0=0$\\
\textbf{for}~$k=1,2,\ldots$ \\
~~~ \textbf{step 1}: ${\bf x}^{k+1}=\textbf{Prox}_{\gamma_k g}[{\bf x}^k-\gamma_k\nabla f({\bf x}^k)+\beta_k({\bf x}^k-{\bf x}^{k-1})]$ \\
~~~ \textbf{step 2}:  update as  \eqref{schemeAup}\\
\textbf{end for}\\
\end{algorithmic}
\end{algorithm}

\begin{algorithm}[H]

\caption{Ada with Adaptive Momentum (Ada$^2$m)}
\begin{algorithmic}\label{alg4}
\REQUIRE   parameters $(\gamma_k)_{k\geq 0}\subseteq(0,+\infty)$, $0<\delta\leq 1$, $(\alpha_k)_{k\geq 0}\subseteq[0,1)$\\
\textbf{Initialization}:   ${\bf x}^0={\bf x}^1=\textbf{0}$, $\beta_0=0$\\
\textbf{for}~$k=1,2,\ldots$ \\
~~~ \textbf{step 1}: ${\bf x}^{k+1}={\bf x}^k-\gamma_k \frac{{\bf m}^k}{[{\bf v}^k]^{\frac{1}{2}}}$ \\
~~~ \textbf{step 2}: get ${\bf g}^k=\sum_{j=1}^m \nabla f_{i_j}({\bf x}^k)/m$\\
~~~ \textbf{step 3}: ${\bf m}^{k+1}=\beta_k{\bf m}^k+(1-\beta_k){\bf g}^k$\\
~~~ \textbf{step 4}: ${\bf v}^{k+1}=\alpha_k {\bf v}^k+(1-\alpha_k)[{\bf g}^k]^2$\\
~~~ \textbf{step 5}:  update as \eqref{schemeAup}\\
\textbf{end for}\\
\end{algorithmic}
\end{algorithm}

\section{Convergence Analysis}\label{sec:analysis}
This part consists of the convergence analysis for our proposed algorithms. First, we collect several necessary assumptions that are widely used in (non)convex stochastic optimization.

\textbf{Assumption 1}: The stochastic gradient is an unbiased estimate, i.e., $\EE {\bf g}^k= \nabla f({\bf x}^k)$.

\textbf{Assumption 2}: The gradient of $f$ is $L$-Lipschitz, i.e., $\|\nabla f({\bf x})-\nabla f({\bf y})\|\leq L\|{\bf x}-{\bf y}\|$ with $L>0$.

\textbf{Assumption 3}: The stochastic gradient is uniformly bounded, i.e., $\sup_{k}\{\|{\bf g}^k\|\}\leq R$ with $R>0$.

\subsection{Convergence of ASHB}
We prove the convergence of ASHB in strongly convex (Theorem~\ref{th2}), general-convex (Theorem~\ref{th3}), and nonconvex (Theorem~\ref{th4}) settings.

\begin{theorem}[Strong convexity case]\label{th2}
Let $f({\bf x})$ be strongly convex and twice-differentiable, and let $\{{\bf x}^k\}_{k\geq 0}$ be generated by \eqref{scheme}. Assume that $0<\nu\leq \min_{{\bf x}}\{\lambda_{\min}(\nabla^2 f({\bf x}))\}\leq \max_{{\bf x}}\{\lambda_{\max}(\nabla^2 f({\bf x}))\}\leq L$, and Assumptions 1, 2, 3 hold.
Then,
$$\EE\|{\bf x}^{K}-{\bf x}^*\|\leq (1-2\gamma\nu)^K\EE\|{\bf x}^{1}-{\bf x}^*\|^2 +C_1\gamma,$$
where $C_1>0$ is a constant independent on $\gamma$, $d$, and $K$.
\end{theorem}
The obtained convergence rate in Theorem~\ref{th2} is similar to SGD for strongly convex optimization, i.e., a linear rate at the beginning. Also, the error term in the above convergence rate needs to be removed by using a diminishing learning rate. Thus, to reach $\epsilon>0$ error for $\EE\|{\bf x}^{K}-{\bf x}^*\|$, we need to set $\gamma=\Theta(\epsilon)$ and $K=\mathcal{O}(\frac{1}{\nu\epsilon}\ln \frac{1}{\epsilon})$.

For general convex cases, the convergence of Algorithm \ref{alg1} can be described as follows.
\begin{theorem}[General convexity case]\label{th3}
Let $f$ be convex and $\{{\bf x}^k\}_{k\geq 1}$ be generated by \eqref{scheme}. Suppose Assumptions 1, 2 and 3 hold. Then
\begin{align*}
\EE f\Big(\sum_{k=1}^K{\bf x}^k/K\Big)-\min f
\leq \EE\|{\bf x}^{1}-{\bf x}^*\|^2/(2\gamma K)+C_2\gamma,
\end{align*}
where $C_2>0$ is a constant independent on $\gamma$, $d,$ and $K$.
\end{theorem}
Given the desired $\epsilon$ error tolerance, we need to set $\gamma=\Theta(\epsilon)$ and let  $K=\mathcal{O}(\frac{1}{\epsilon^2})$.
For general nonconvex optimization problems, we measure the convergence of gradient norm following current routines. We have the following convergence guarantee for ASHB for nonconvex optimization.
\begin{theorem}[Nonconvexity case]\label{th4}
Let $\{{\bf x}^k\}_{k\geq 1}$ be generated by \eqref{scheme} and suppose Assumptions 1, 2 and 3 hold. It follows that
\begin{align*}
\min_{1\leq k\leq K}\{\EE\|\nabla f({\bf x}^{k})\|^2\}\leq C_3\gamma
+\frac{C_4(f({\bf x}^1)-\min f)}{\gamma K},
\end{align*}
where $C_3,C_4>0$ are constants independent on $\gamma$, $d$, and $K$
\end{theorem}
To get the $\epsilon$ error for $\min_{1\leq k\leq K}\{\EE\|\nabla f({\bf x}^{k})\|^2\}$, we need to set $\gamma=\Theta(\epsilon)$ and  $K=\mathcal{O}(\frac{1}{\epsilon^2})$.
In summary, we cannot prove the theoretical acceleration of the proposed adaptive momentum algorithms. But the obtained results above show that the speed of ASHB can run 
as fast as SGD in strongly convex, general-convex, and nonconvex cases. Also, the convergence guarantees can promise the use of the proposed SGD with adaptive momentum, which is better than the stochastic Nesterov's acceleration and its restart variants.


\subsection{Convergence of PAHB}
In this subsection, we analyze the convergence of PAHB; we summarize our theoretical convergence guarantee for PAHB in the following proposition.
\begin{proposition}[Convergence of PAHB]\label{proprox}

Assume that $\{{\bf x}^k\}_{k\geq 1}$ is generated by PAHB. If $g$ is  nonconvex, and $\gamma_k=\frac{c(1-2\beta_k)}{L}$ with $0<c<1$, and $\frac{1}{2}<\delta\leq 1$. Then we have
$\sum_{k}\|{\bf x}^{k+1}-{\bf x}^k\|^2<+\infty.$
If $g$ is convex, the result still holds when  $\gamma_k=\frac{c(1-\beta_k)}{L}$ with $0<c<1$, and $0<\delta\leq 1$.
\end{proposition}
\subsection{Convergence of Ada$^2$m}
In this subsection, we discuss the convergence of Ada$^2$m for nonconvex optimization. We summarize 
the convergence of Ada$^2$m in the following proposition.
\begin{proposition}[Convergence of Ada$^2$m]\label{proadam}
Let $\{{\bf x}^k\}_{k\geq 1}$ be generated by Ada$^2$m. Let $\alpha_k=\frac{1}{k}$ and $\gamma_k=\frac{\gamma}{\sqrt{k}}$ for some $\gamma>0$, and $\|{\bf g}^1\|\geq \pi>0$. Then, 
$$\min_{1\leq k\leq K}\{\EE\|\nabla f({\bf x}^k)\|^2\}=\mathcal{O}(\frac{\ln K}{\sqrt{K}}).$$
\end{proposition}
We provide technical proofs for all the above theoretical results in the appendix.

\section{Numerical Results}\label{sec:numerical}
In this section, we verify the efficiency of the proposed adaptive momentum for training various machine learning models, including regularized logistic regression models, ResNets, LSTMs, and transformers.
In all the following experiments, we show the advantage of the adaptive momentum \eqref{eq:adp:momentum} over the baseline algorithms with a well-calibrated momentum. All experiments are conducted on a server with 4 NVIDIA 2080TI GPUs.
\begin{figure*}[!ht]
\centering
\begin{tabular}{cccc}
\hskip -0.3cm\includegraphics[width=0.25\linewidth]{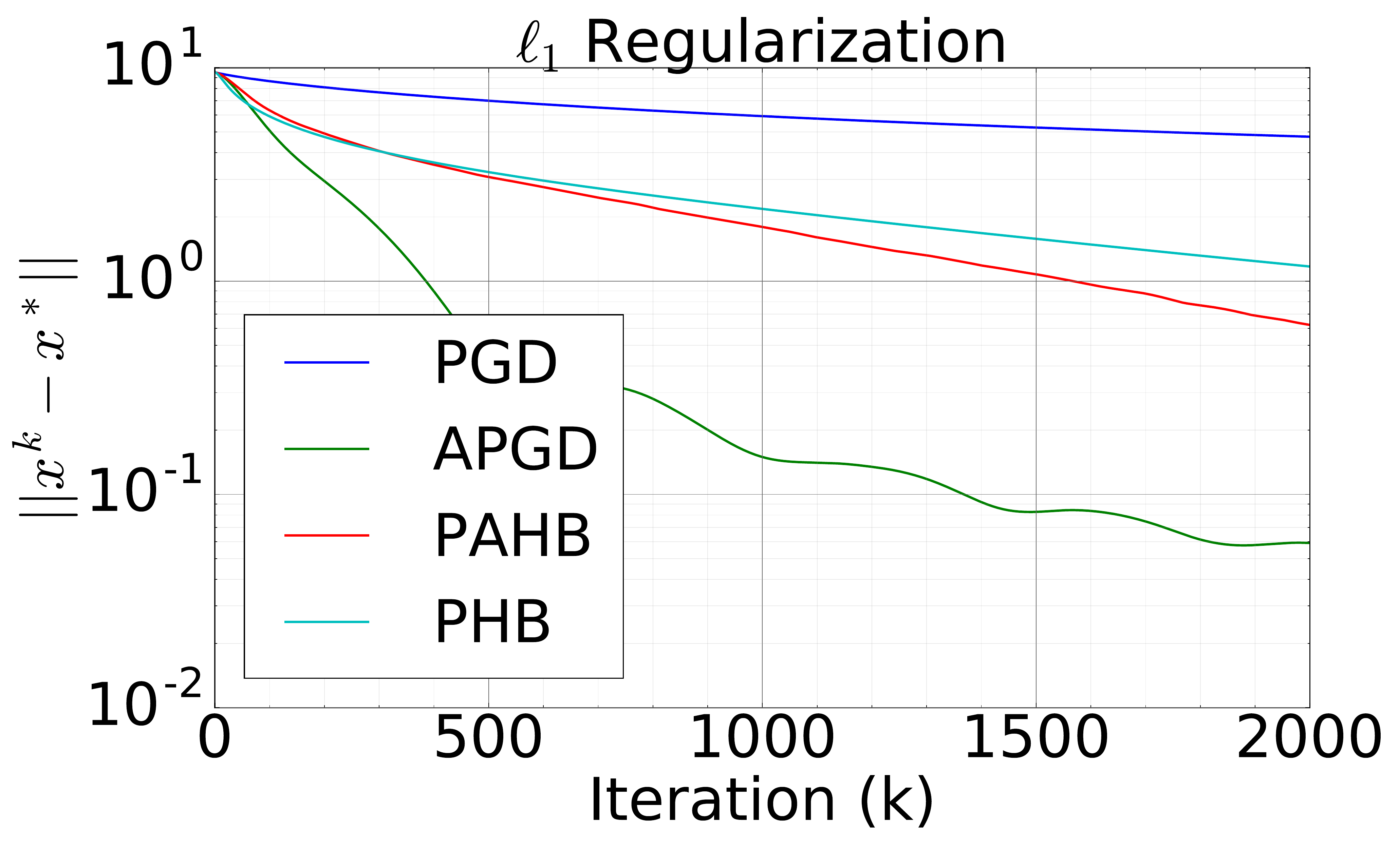}&
\hskip -0.4cm\includegraphics[width=0.25\linewidth]{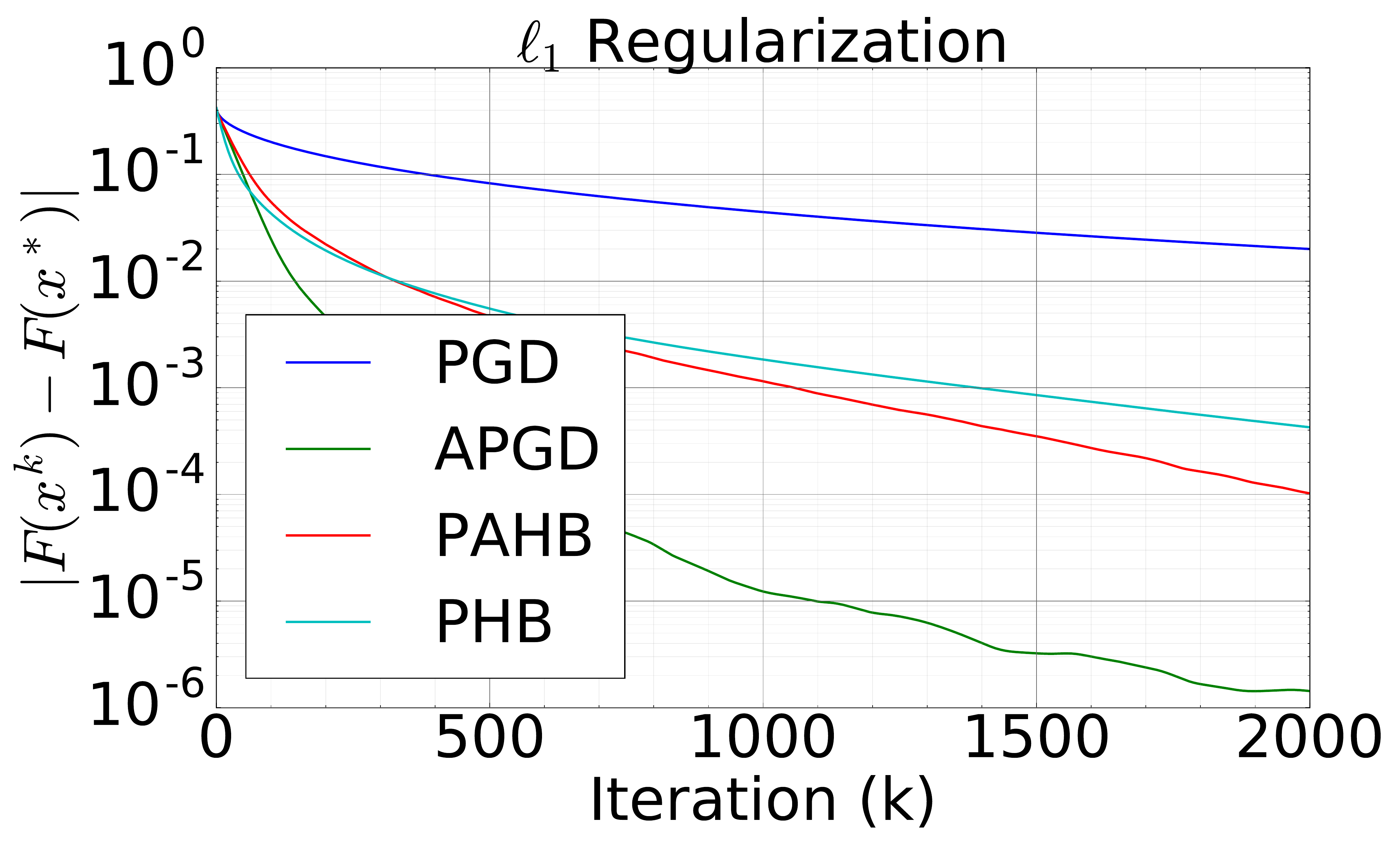}&
\hskip -0.4cm\includegraphics[width=0.25\linewidth]{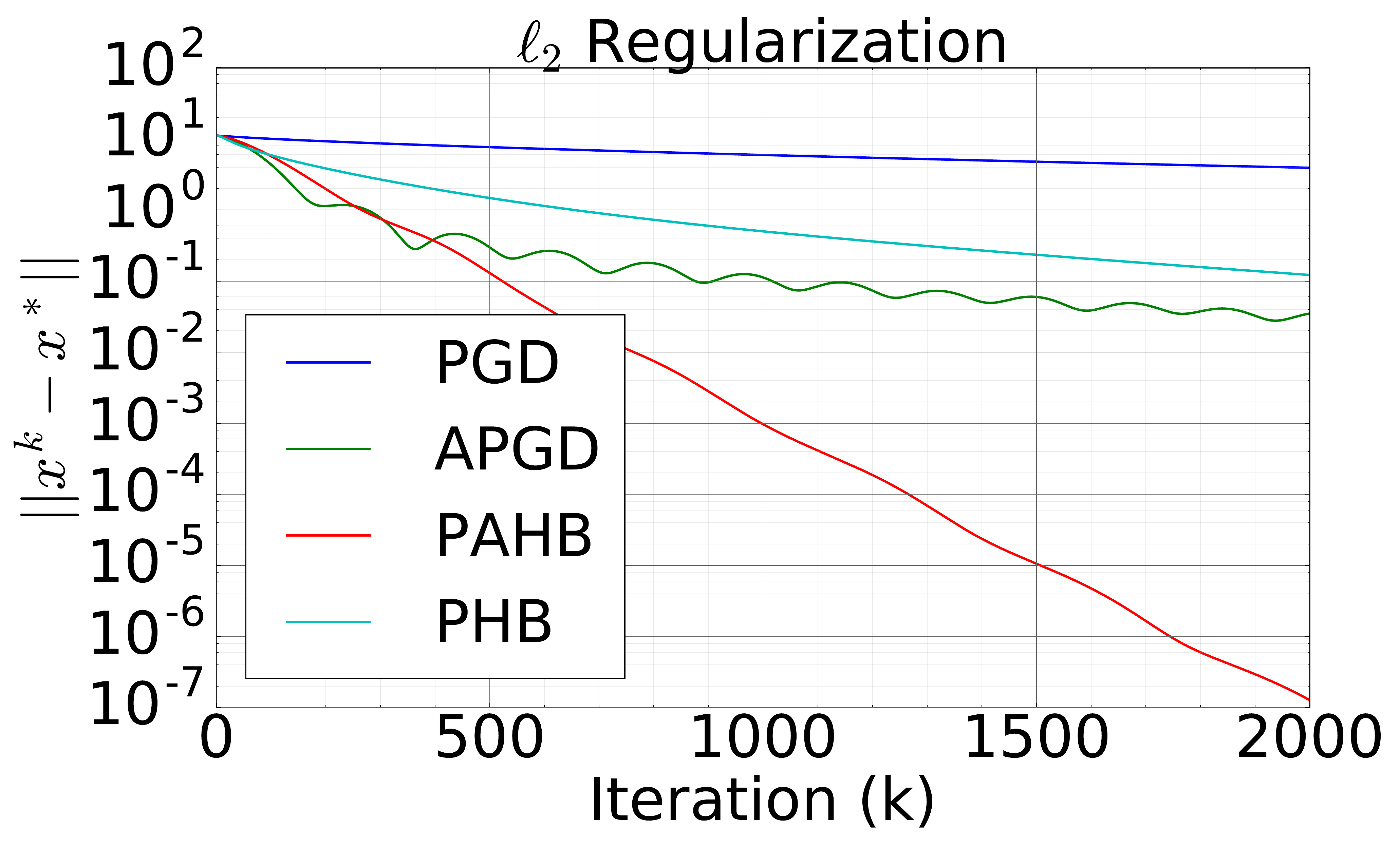}&
\hskip -0.4cm\includegraphics[width=0.25\linewidth]{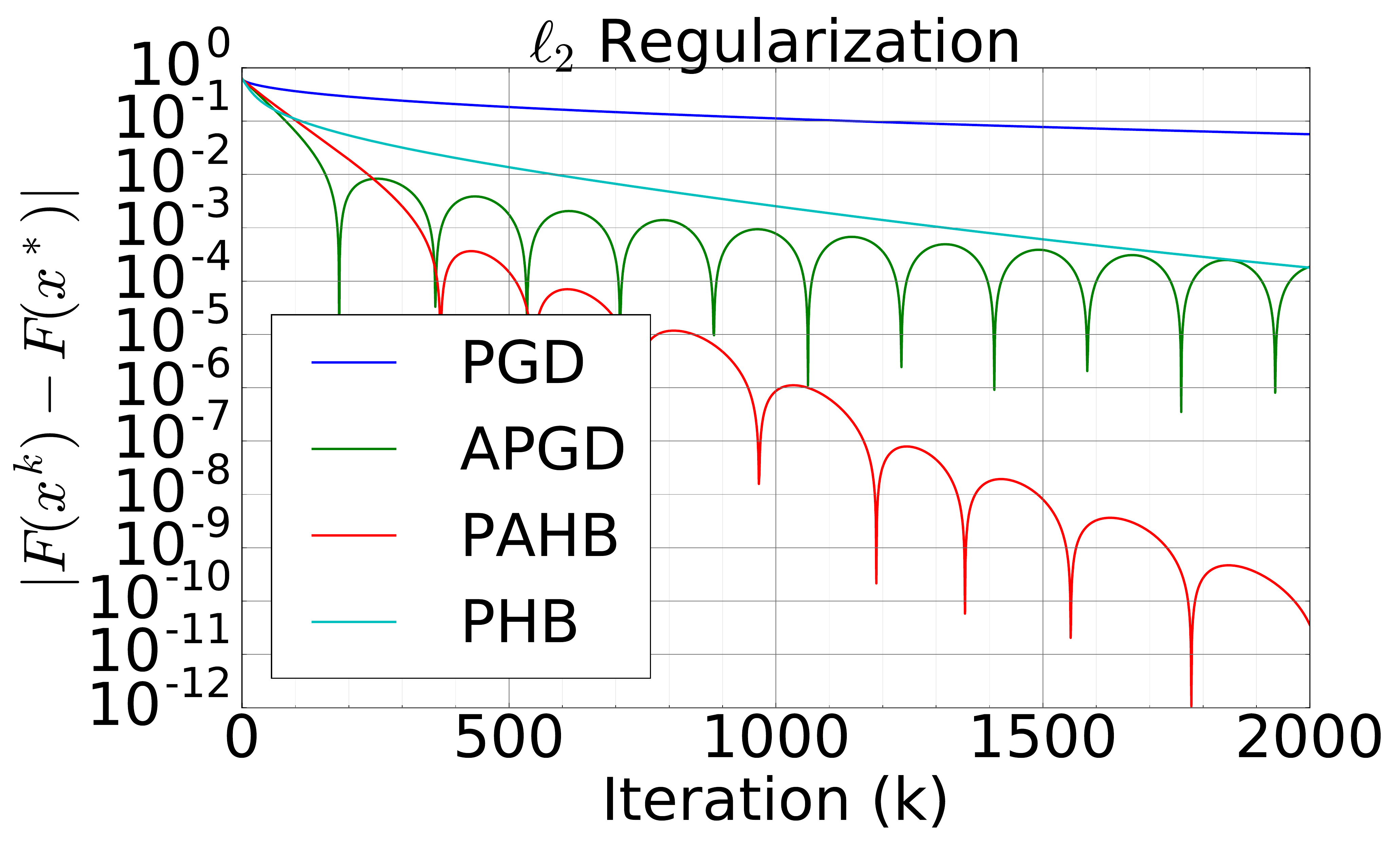}\\
\end{tabular}
\caption{Iteration ($k$) vs. $\|{\bf x}^k-{\bf x}^*\|$ and $|F({\bf x}^k)-F({\bf x}^*)|$, which are obtained by using different optimization algorithms for solving regularized Logistic regression problem. PAHB outperforms PHB with both $\ell_1$- and $\ell_2$-regularizers; PAHB is the fastest algorithm when the $\ell_2$ regularization is used. We obtain the minimum ${\bf x}^*$ for both regularizers by running PGD for $10^6$ iterations.
}
\label{fig:prox}
\end{figure*}
\subsection{Regularized Logistic Regression}
Consider training the following regularized logistic regression model
\begin{equation}\label{eq:prox-problem}
\min_{\bf x}\{F({\bf x}) := f({\bf x}) + \lambda R({\bf x})\},
\end{equation}
where $f({\bf x})=\frac{1}{2n}\sum_{i=1}^n\log(1+\exp(-b_i{\bf a}_i^T{\bf x}))$ with $n$ be the sample size and ${\bf a}_i\in \mathbb{R}^d$ is a training data,  
$b_i\in\{-1,1\}$ be the label of ${\bf a}_i$, and $\lambda=0.001$.
Here, we consider both $\ell_1$ ($R({\bf z})=\|{\bf z}\|_1$) and $\ell_2$ ($R({\bf z})=\|{\bf z}\|_2^2$) regularization.

For \eqref{eq:prox-problem}, we compute the exact gradient of $f({\bf x})$ and compare PAHB with three well-studied benchmark algorithms, namely, proximal gradient descent (PGD), accelerated proximal gradient descent (APGD)~\cite{tibshirani2010proximal}, and proximal heavy ball (PHB) (replace $\beta_k$ with a constant $\beta$ in PAHB).
We set $n=d=1000$ and let $[{\bf a}_1,{\bf a}_2,\cdots,{\bf a}_n]$ be the $n\times d$ multivariate-normal matrix with the covariance matrix being Toeplitz, whose first row is $[0.9^0, 0.9^1,\cdots]$. ${\bf x}\in \mathbb{R}^{1000}$ is a vector whose $i$-th entry is $(-1)^ie^{-i/100}$ for $0\leq i\leq 500$ and $0$ otherwise. $b_i$ is a binomial random variable with probability ${\bf a}^T{\bf b}$ to take the value $1$. Figure~\ref{fig:prox} shows the comparisons between the above four algorithms, where we begin with the zero initialization and use a learning rate of $0.1$ for the above four algorithms. For PHB, we set $\beta=0.9$, which is obtained by grid search. It is seen that PHB is faster than PGD, and adaptive momentum can remarkably accelerate PHB. Moreover, PAHB is the fastest solver for solving the $\ell_2$-regularized logistic regression problem among the above four algorithms.

\subsection{Training DNNs for Image Classification}
\subsubsection{ResNets for CIFAR10 Classification}
\begin{table}[H]
\caption{Test accuracy (\%) of PreResNet20/56/110 for CIFAR10 classification, where the models are trained by the Adam-style algorithms. (five independent runs).
}
\label{tab:acc:cifar10:adam}
\centering
\vspace{2mm}
\fontsize{8.5pt}{0.85em}\selectfont
\begin{tabular}{c|c|c|c}
\hline
Model & PreResNet20 & PreResNet56 & PreResNet110\\
\hline
Adam      & $91.15\pm 0.06$  & $92.36\pm 0.09$ & $93.17\pm 0.10$\\
Ada$^2$m  & $91.24\pm 0.07$  & $92.65\pm 0.07$ & $93.31\pm 0.09$\\
AdamW     & $90.65\pm 0.09$  & $92.64\pm 0.13$ & $93.46\pm 0.15$\\
Ada$^2$mW & $90.99\pm 0.08$  & $92.80\pm 0.04$ & $93.65\pm 0.07$\\
\hline
\end{tabular}
\end{table}

\begin{table}[H]
\caption{Test accuracy (\%) of PreResNet20/56/110 for CIFAR10 classification, where the models are trained by SGDM and ASHB algorithms. (five independent runs).} \label{tab:acc:cifar10:sgdm}
\centering
\vspace{2mm}
\fontsize{8.5pt}{0.85em}\selectfont
\begin{tabular}{c|c|c|c}
\hline
Model & PreResNet20& PreResNet56& PreResNet110\\
\hline
SGDM & $92.35 \pm 0.09$& $93.62 \pm 0.11$ & $94.75 \pm 0.14$\\
ASHB & $92.48\pm 0.04$ & $93.99 \pm 0.06$ & $95.12 \pm 0.09$\\
\hline
\end{tabular}
\end{table}

We train ResNet20/56/110 with pre-activation \cite{he2016identity}, denoted as PreResNet20/56/110, using batch size 128 for CIFAR10 classification, which contains 50K/10K images in the training/test set. 

\begin{figure}
\begin{tabular}{cc}
\hskip -0.3cm\includegraphics[width=0.24\textwidth]{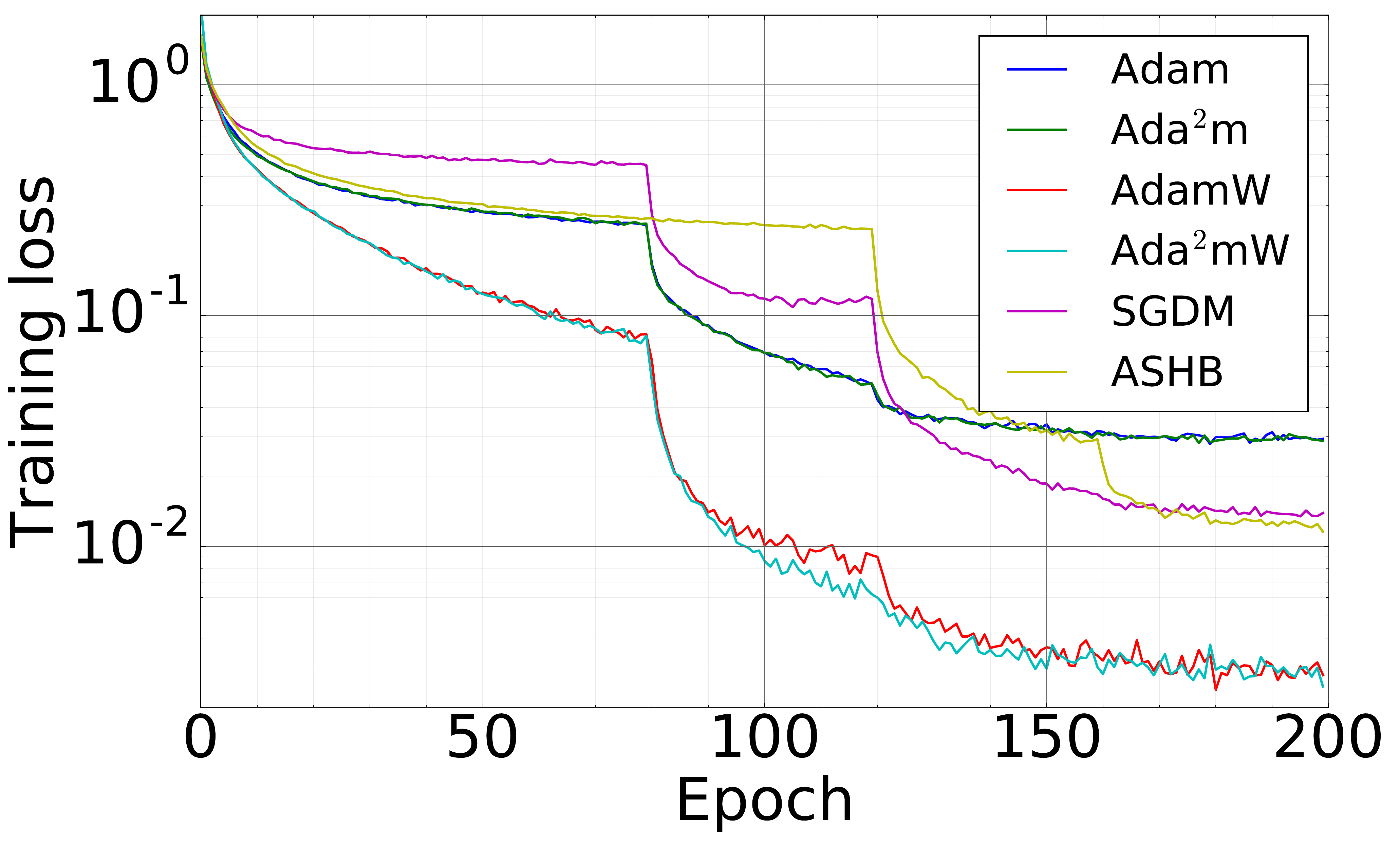}&
\hskip -0.3cm\includegraphics[width=0.24\textwidth]{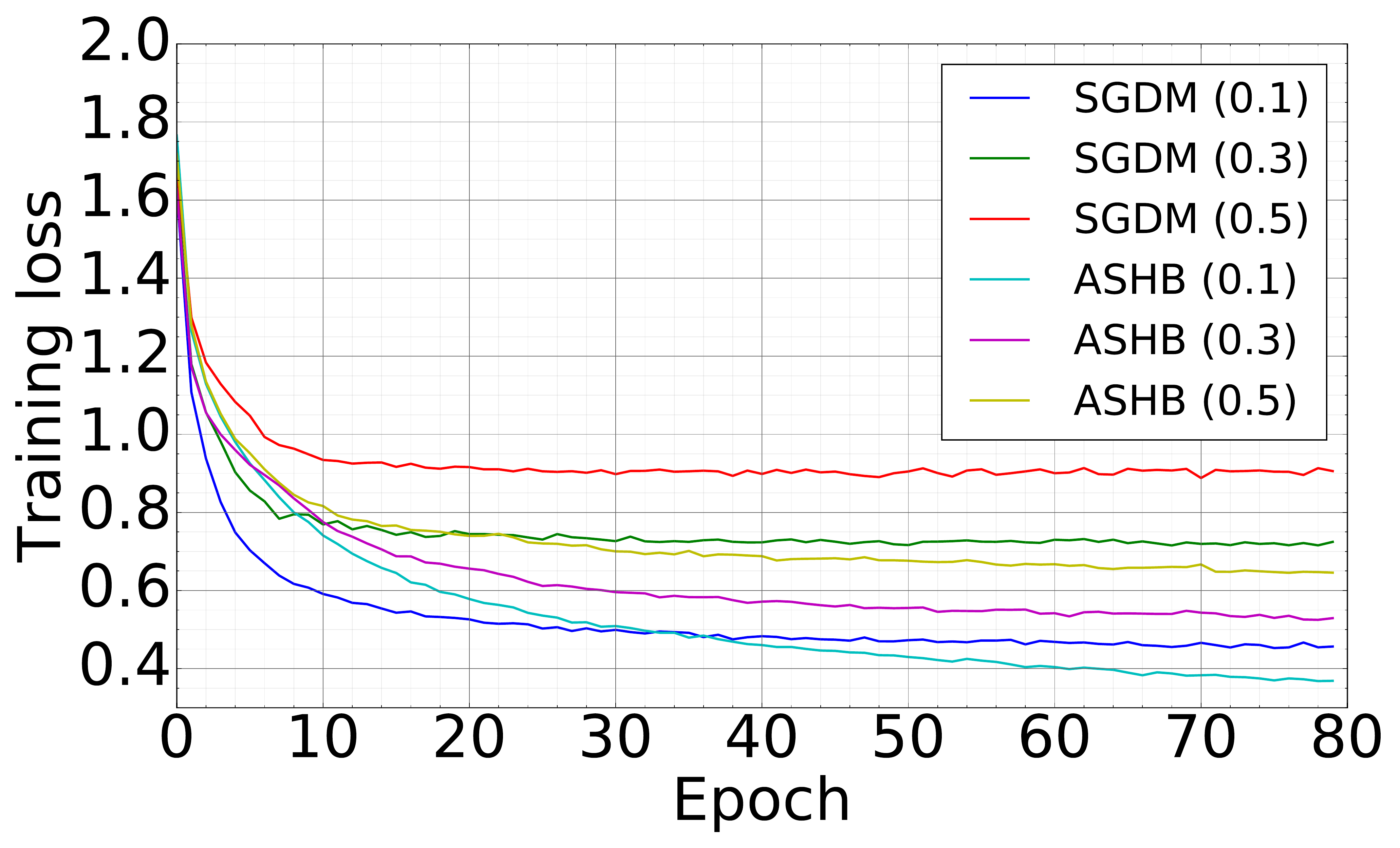}\\
\end{tabular}
\caption{Left: Epoch vs. training loss obtained from training PreResNet56 for CIFAR10 classification by using different optimizers. Right: Comparisons between SGDM and ASHB under different learning rates. ASHB converges faster than SGDM under a given learning rate (in parenthesis).
Also, ASHB is more robust to larger learning rates
than SGDM.
}
\label{fig:training-curve}
\end{figure}

\paragraph{Adam-style algorithms} We consider two popular Adam-style algorithms that have been implemented in the latest version of PyTorch \cite{NEURIPS2019_9015}, namely the Adam and AdamW\cite{loshchilov2017decoupled}, for training ResNets\footnote{We can also incorporate the adaptive momentum with other Adam variants.}. We run both algorithms for 200 epochs using an initial learning rate of 0.001/0.003 for Adam/AdamW \footnote{We perform grid search for the initial learning rate to obtain this choice for Adam and AdamW, respectively.} and decay the learning rate by a factor of 10 at the 80-th, 120-th, and 160-th epoch, respectively. We set the weight decay to be $0.0005$. Meanwhile, we train ResNets by Ada$^2$m and Ada$^2$mW using the same setting as that used in Adam and AdamW
\footnote{Note that the learning rate is optimized for Adam and AdamW rather than for Ada$^2$m and Adam$^2$W.}. Table~\ref{tab:acc:cifar10:adam} lists the test accuracy of both models for classifying CIFAR10. We see that adaptive momentum can consistently improve test accuracy for all the three ResNet models. Figure~\ref{fig:training-curve} (Left) plots training curves of the four Adam-style algorithms in training PreResNet56 for CIFAR10 classification. We see that \emph{AdamW and Adam$^2$W converge remarkably faster than Adam and Ada$^2$m, and Ada$^2$mW is even slightly faster than AdamW.}

\paragraph{SGDM vs. ASHB} Now, we conduct the above CIFAR10 training by using SGDM and ASHB. First, we show that ASHB is more robust to different learning rates. Figure~\ref{fig:training-curve} (Right) shows the training curves of SGDM ($\beta=0.9$) and ASHB using different learning rates for training PreResNet20. These results show that with the same learning rate ASHB converges faster than SGDM and ASHB is more resilient to larger learning rates than SGDM. For instance, when the learning rate is 
0.5, the training loss of ASHB keeps decaying, while the training loss of SGDM plateaus after 
15 epochs. \emph{Adaptive momentum enables faster convergence and is more robust to different learning rates.}

Second, we show that ASHB can improve the test accuracy of the trained models. We train the above three ResNet models using SGDM and ASHB for 200 epochs, respectively. For SGDM, we use the benchmark initial learning rate of 0.1 and momentum of 0.9, and we decay the learning rate by a factor of 10 at the 80-th, 120-th, and 160-th epochs, respectively. For ASHB, we use an initial learning rate of 0.2 with the same learning rate decay schedule as that used for SGDM. Table~\ref{tab:acc:cifar10:sgdm} lists the test accuracy of the above three models trained by two different optimizers. \emph{ASHB consistently outperforms SGDM in classification accuracy, and the improvement becomes more significant as the network goes deeper.} Furthermore, the variance of testing accuracies of the models trained by ASHB among different runs is smaller than that of SGDM.

\paragraph{Visualize $\beta_k$}
\begin{figure}
\vspace{-0.5cm}
\begin{center}
\begin{tabular}{cc}
\hskip -0.3cm\includegraphics[width=0.5\linewidth]{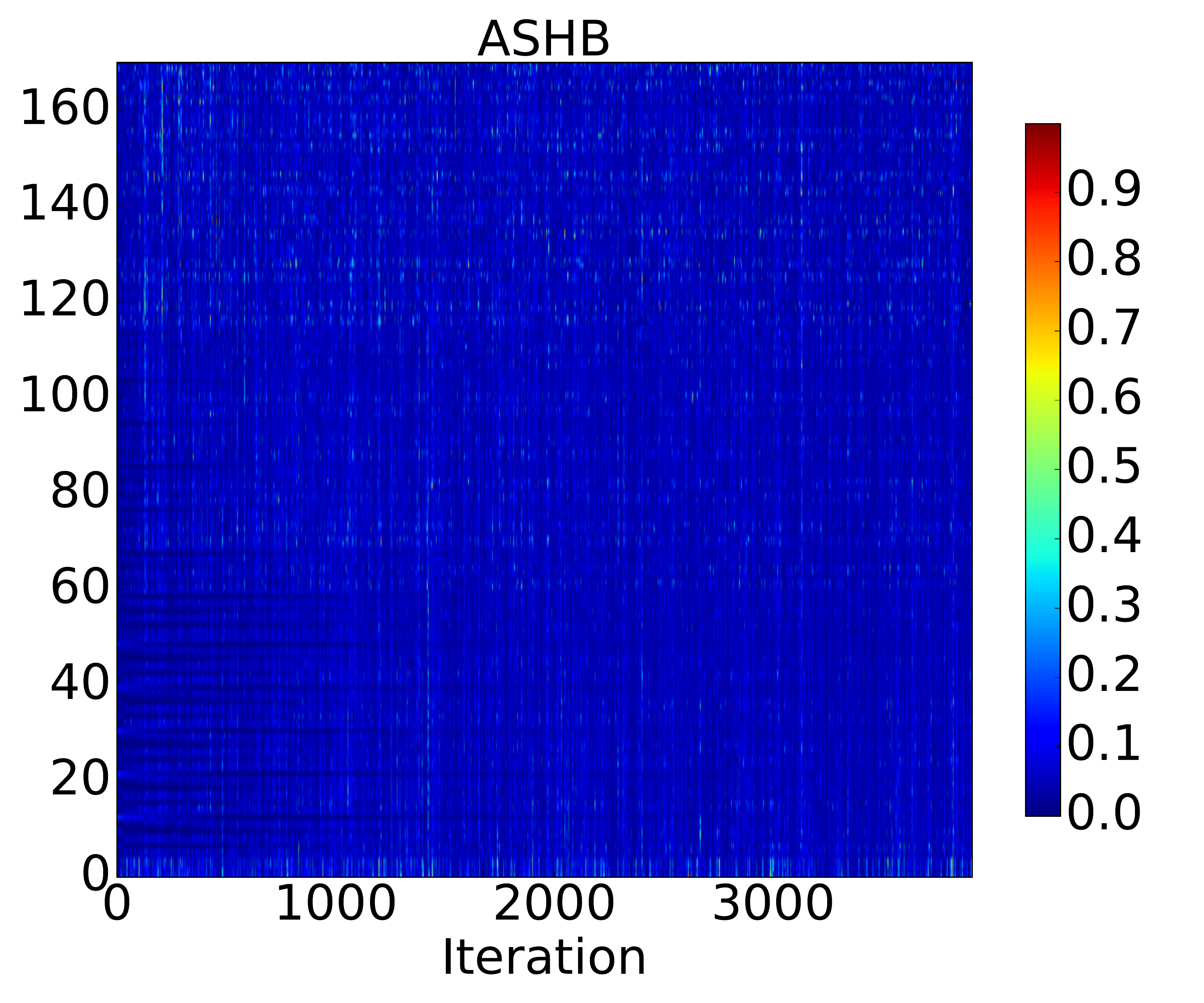}&
\hskip -0.3cm\includegraphics[width=0.5\linewidth]{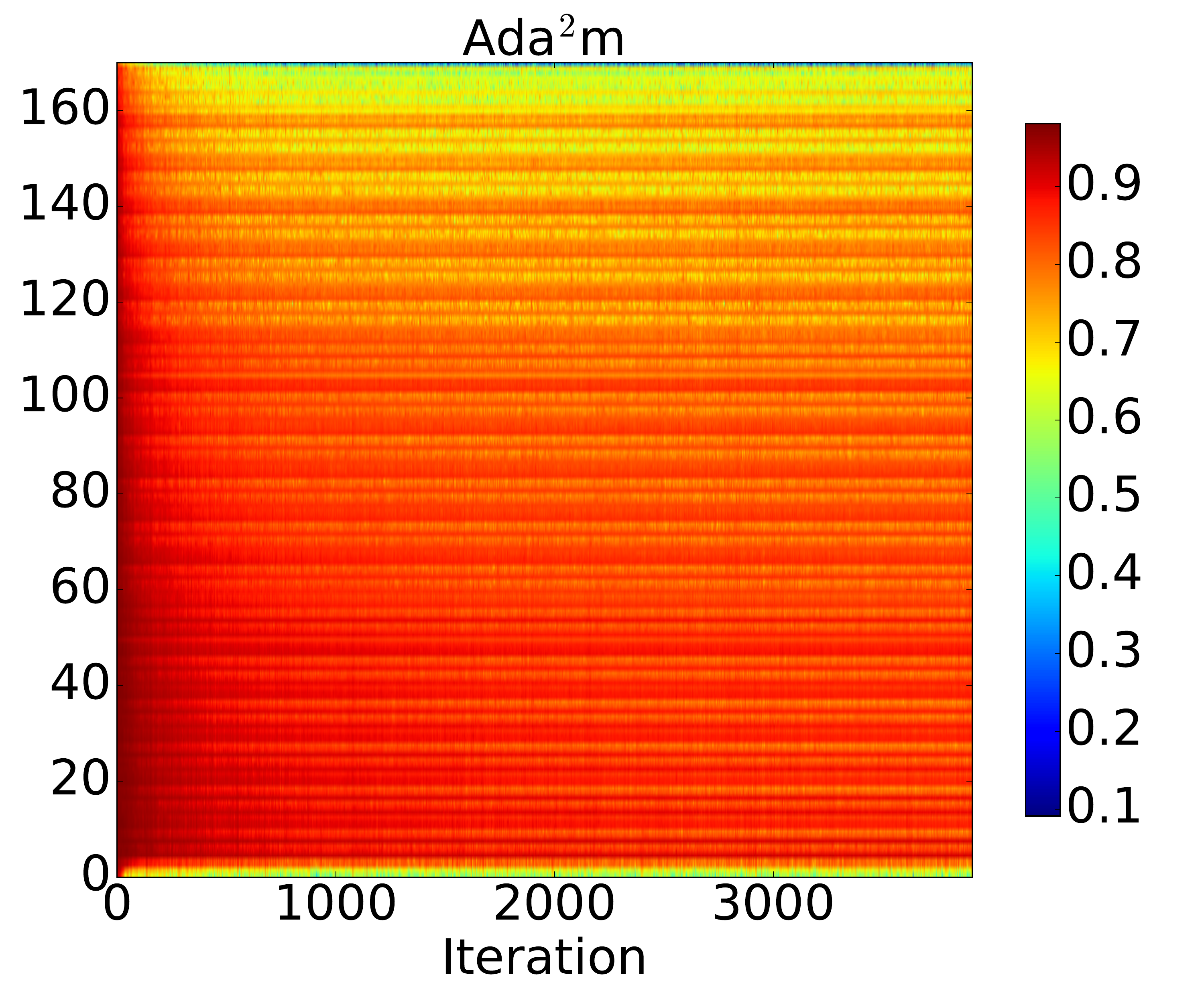}\\[-3pt]
  \end{tabular}
  \end{center}
  \vskip -0.2in
  \caption{Visualization of the evolution of $\beta_k$ for different groups of parameters (170 groups in total) of PreResNet56. In general, the adaptive momentum $\beta_k$ for ASHB is smaller than that of Ada$^2$m.}\label{fig:vis-beta}
\end{figure}\vspace{-0.01in}

 We plot the value of $\beta_k$ for training PreResNet56 using ASHB (initial learning rate 0.2) and Ada$^2$m (initial learning rate 0.001) in Fig.~\ref{fig:vis-beta}. The parameters of PreResNet56 were grouped into 170 groups and the gradient are computed separately via backpropagation, and resulting in 170 different $\beta_k$ values at each iteration $k$. Fig.~\ref{fig:vis-beta} shows that the momentum parameters of ASHB tends to be smaller than that of Ada$^2$m, and different group of parameters using quite different $\beta_k$ when Ada$^2$m is used for training PreResNet56.

\begin{figure}
\renewcommand\arraystretch{1.3}
\setlength{\tabcolsep}{4pt}
\begin{table}[H]
\caption{Test accuracy (\%) of ResNet18 for ImageNet classification, where the models are trained by SGDM (learning rate 0.1) and ASHB (learning rate 0.5). (five independent runs).}
\label{tab:acc:imagenet}
\vspace{2mm}
\centering
\fontsize{8.5pt}{0.85em}\selectfont
\begin{tabular}{c|c}
\hline
Model & ResNet18\\
\hline
SGDM (top-1) & {$69.86\pm 0.048$} {(69.86, \cite{liu2020on})} \\
SGDM (top-5) & {$89.31\pm 0.090$}  \\
ASHB (top-1) & {$69.91\pm 0.057$}  \\
ASHB (top-5) & {$89.36\pm 0.088$}  \\
\hline
\end{tabular}
\end{table}
\end{figure}

\subsubsection{Vision Transformer for CIFAR10 Classification}
\begin{figure*}[ht!]
\centering
\begin{tabular}{cccc}
\includegraphics[width=0.22\linewidth]{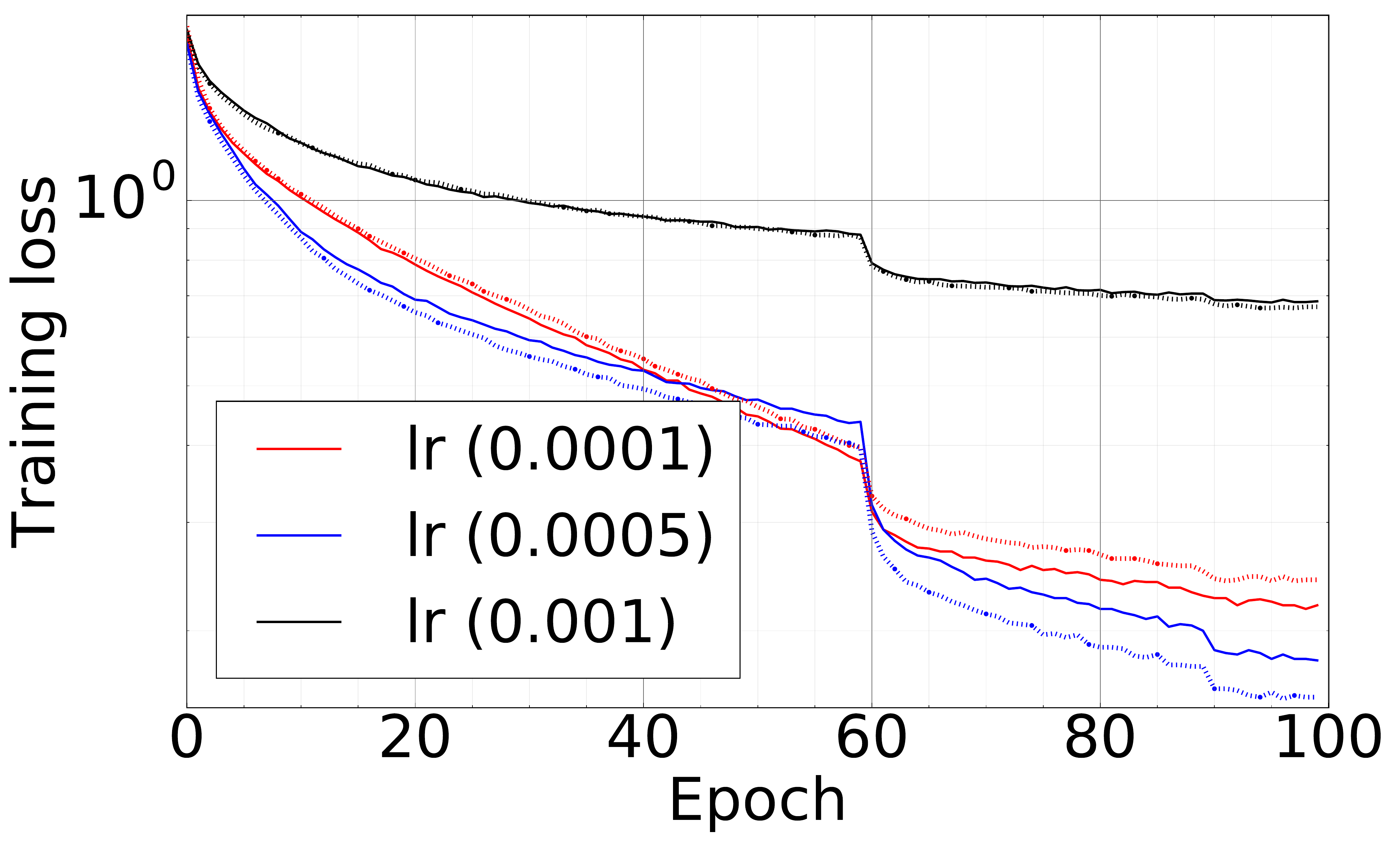}&
\includegraphics[width=0.22\linewidth]{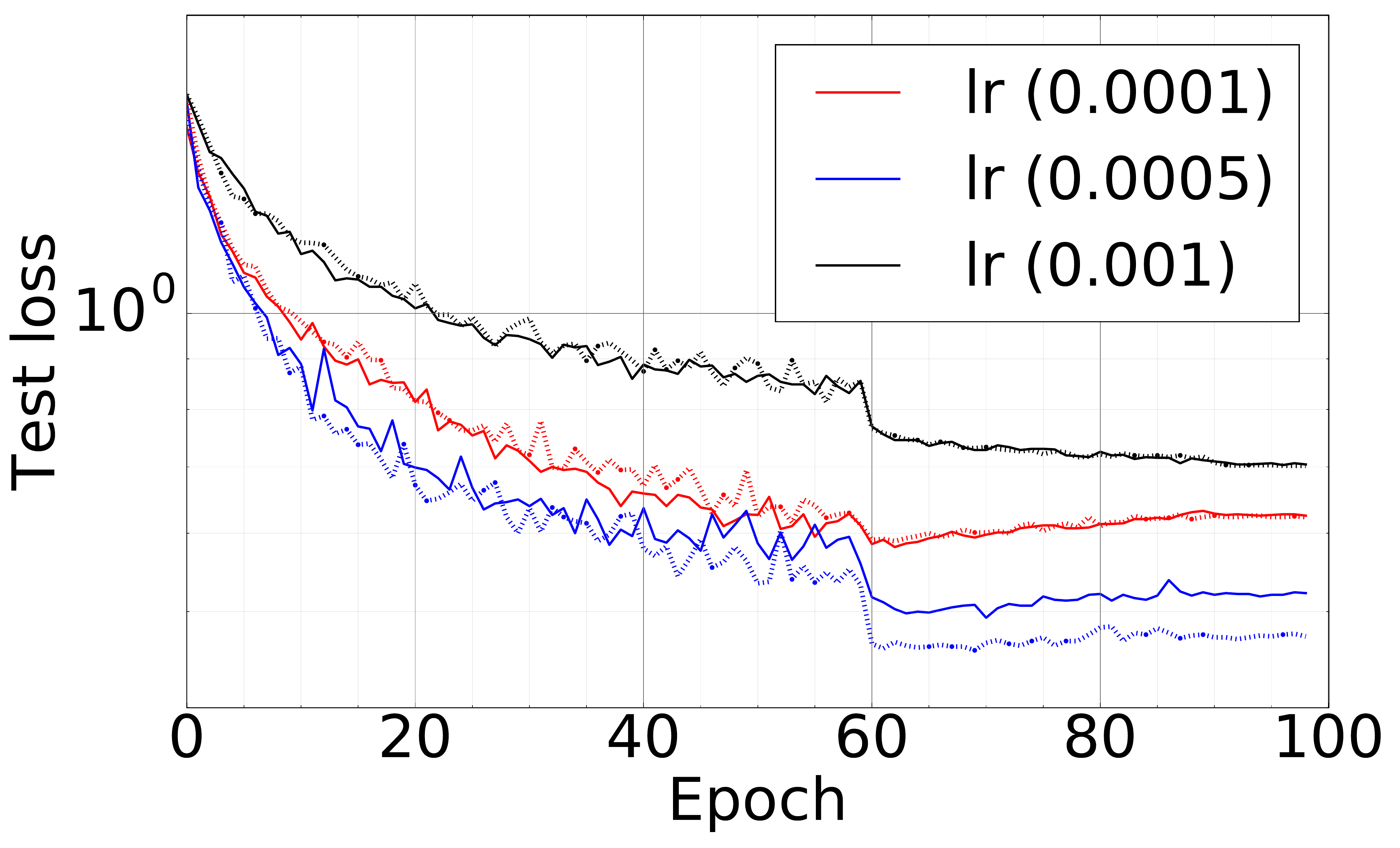}&
\includegraphics[width=0.22\linewidth]{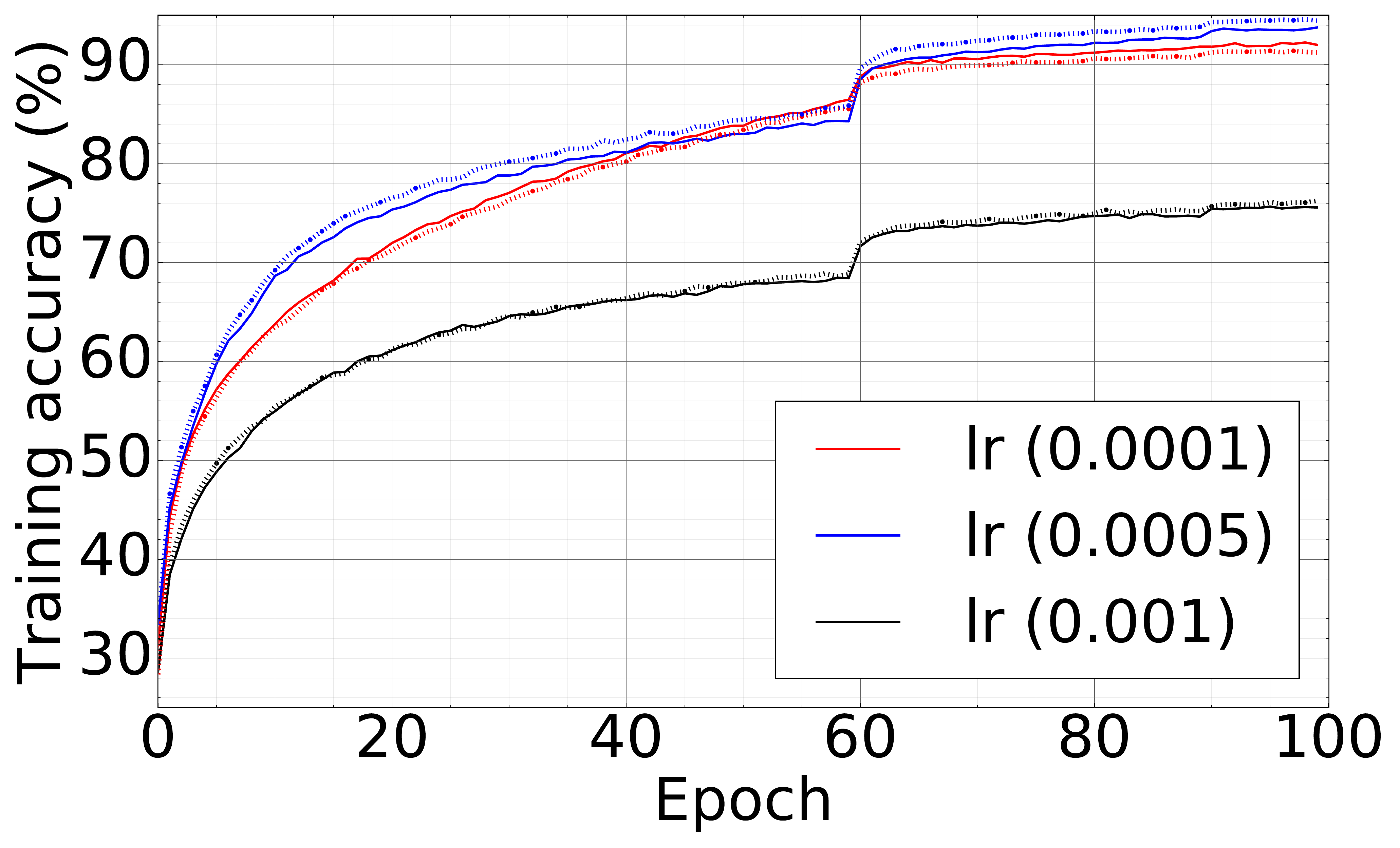}&
\includegraphics[width=0.22\linewidth]{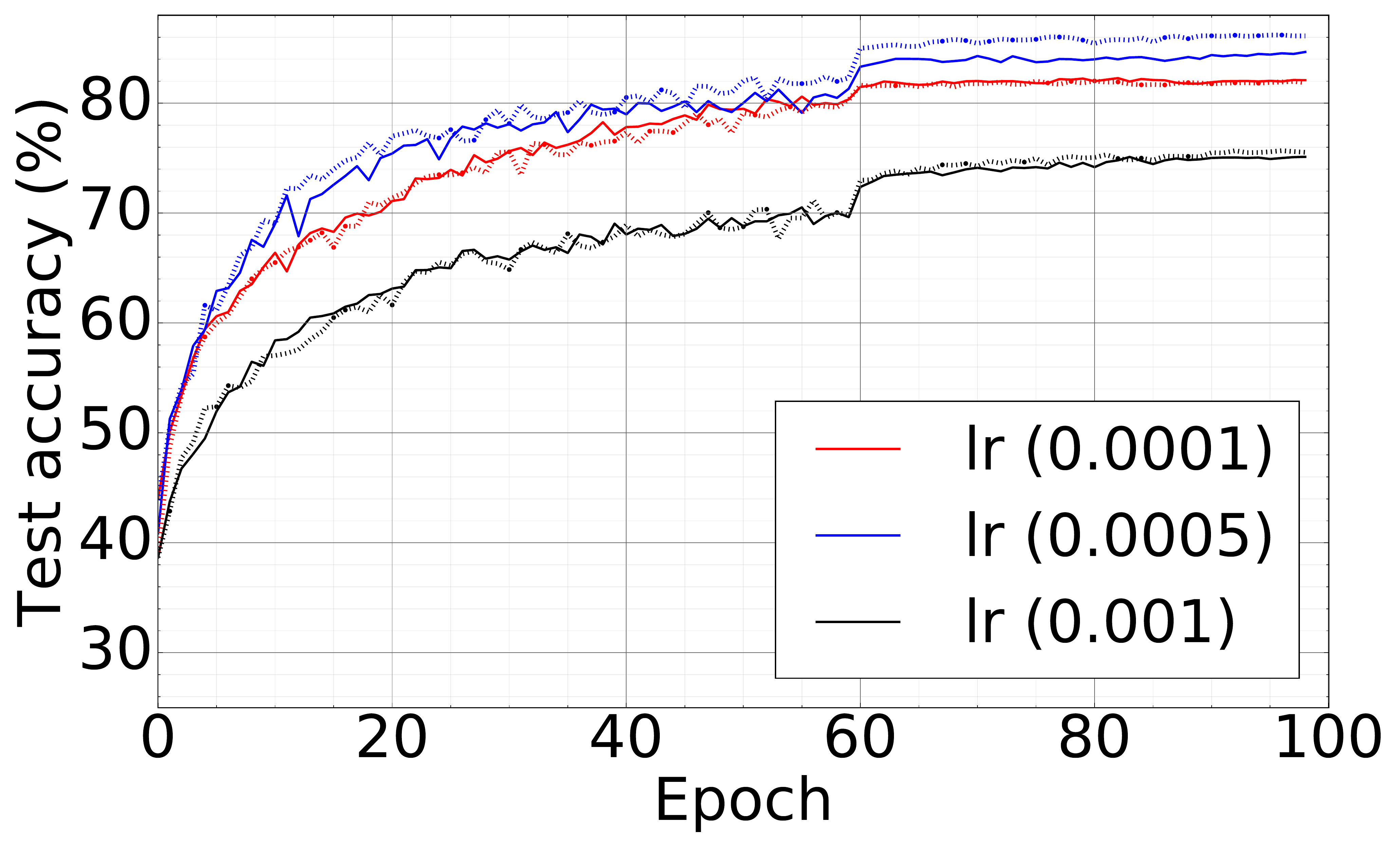}\\
\end{tabular}
\vskip -0.4cm
\caption{Training vision transformer with Adam and Ada$^2$m using different learning rates, solid lines: Adam; dotted lines: Ada$^2$m. Adam and Ada$^2$m perform best when the learning rate (lr) is set to be 0.0005, in which case Ada$^2$m remarkably outperforms Adam.}
\label{fig:vit}
\end{figure*}

We further train the recently developed vision transformer \cite{dosovitskiy2020image} from scratch by using the default Adam algorithm and Ada$^2$m for CIFAR10 classification. We employ the existing PyTorch implementation of the vision transformer \cite{vitgithub} with the same setting, except that we reduce the learning rate by a factor of 10 at the 60-th and 80-th epoch, respectively.
The default learning rate for Adam, in this case, is $10^{-4}$, and we test three different learning rates for both Adam and Ada$^2$m, namely, $10^{-4}, 5\times 10^{-4}$, and $10^{-3}$. Figure \ref{fig:vit} plots the training and test loss and accuracy curves for different settings, and these results show that both Adam and Ada$^2$m perform best when the learning rate is set to be 0.0005, in which case \emph{Ada$^2$m remarkably outperforms Adam in both convergence speed and test accuracy.}

\subsection{ResNets for ImageNet Classification}
\begin{figure}
\begin{center}
\begin{tabular}{cc}
\hskip-0.3cm\includegraphics[width=0.5\linewidth]{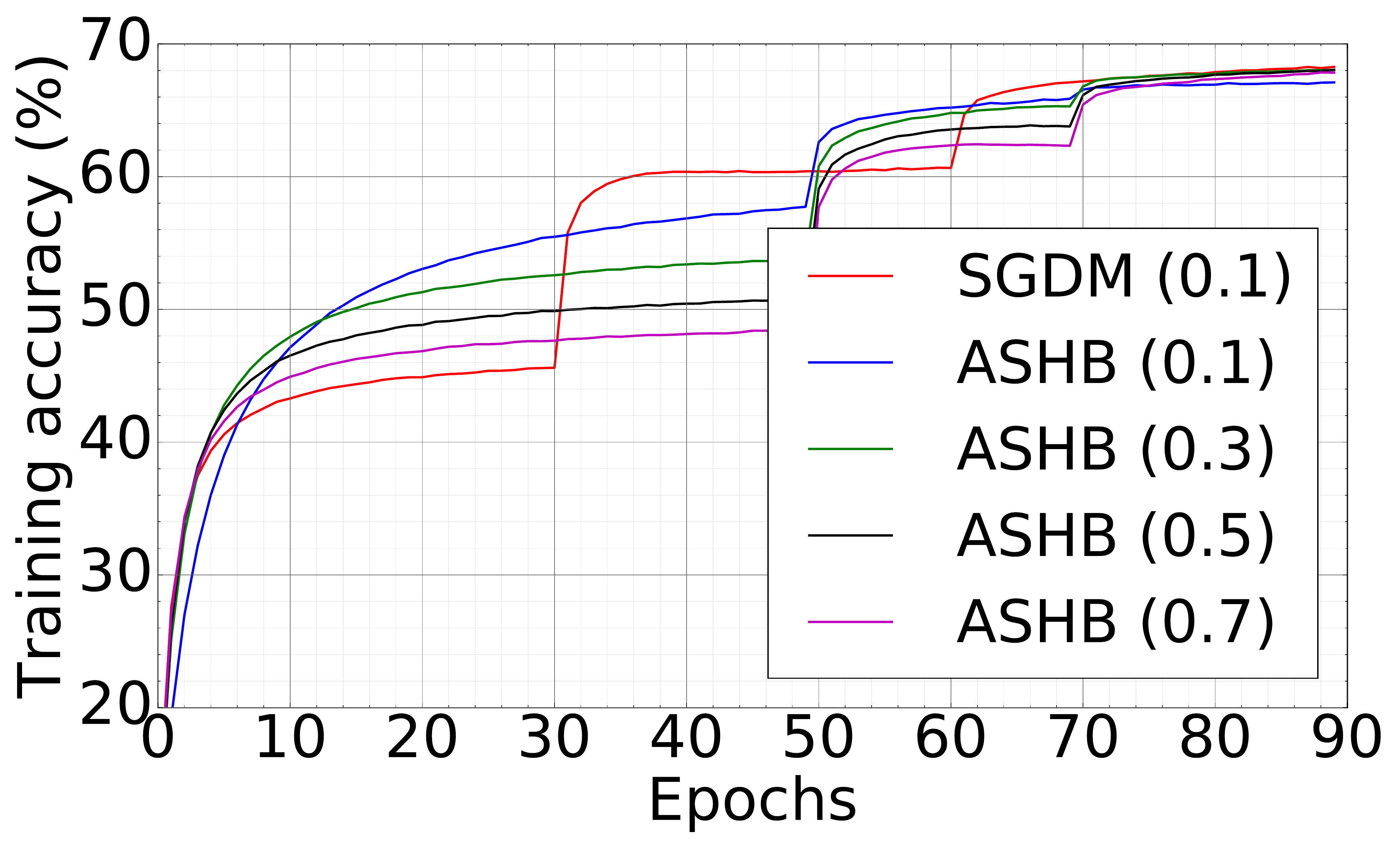}&
\hskip-0.3cm\includegraphics[width=0.5\linewidth]{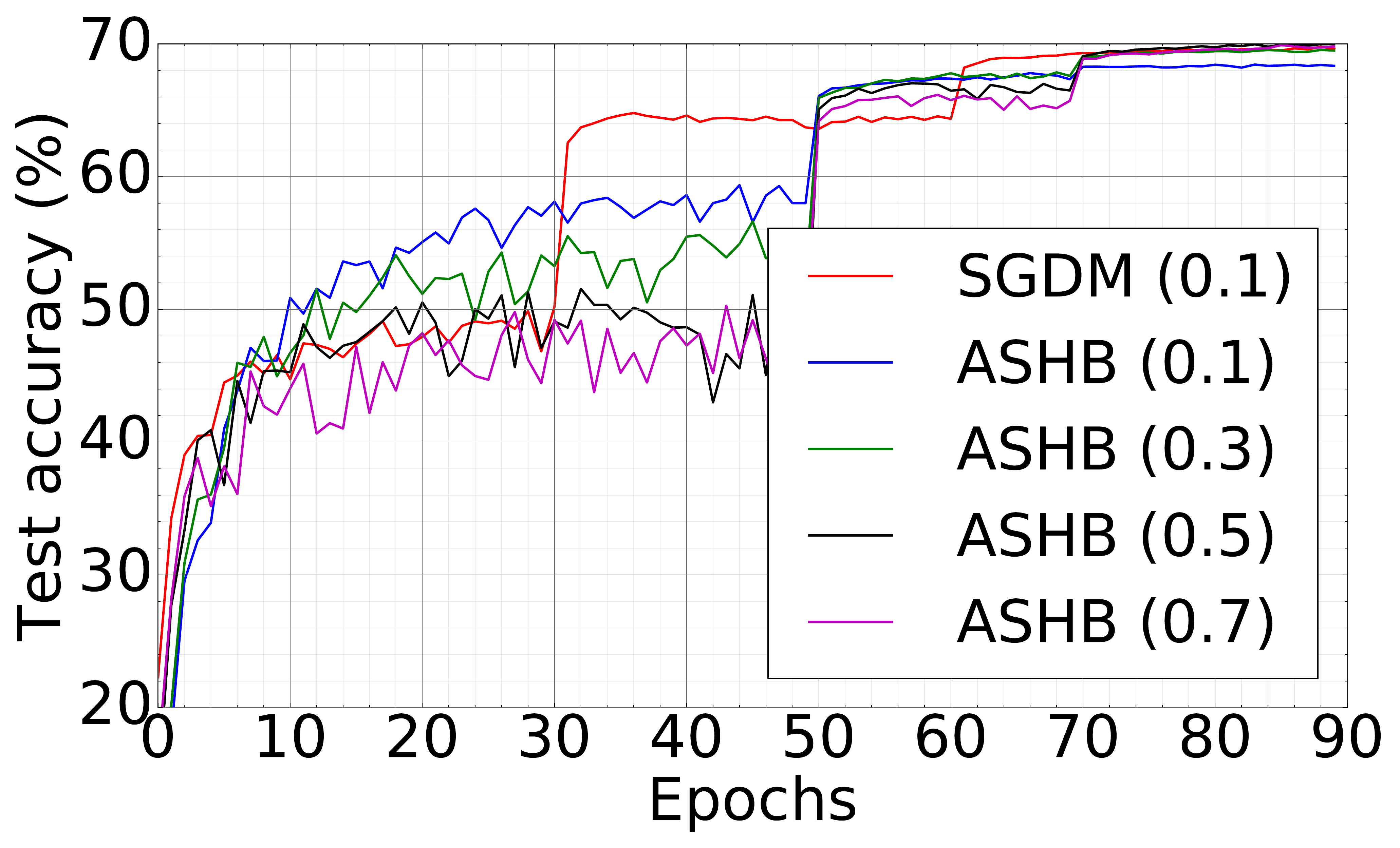}\\[-3pt]
  \end{tabular}
  \end{center}
  \caption{Epochs vs. (top-1) training and 
  test accuracy of ResNet18 trained by SGDM and ASHB, using different learning rates (in parenthesis), for ImageNet classification. In the first learning rate stage, we see that HBAdp is faster than SGDM  using the same learning rate of 0.1. Furthermore, with an appropriate learning rate, e.g., 0.5 and 0.7, the final testing accuracy of the model trained by HBAdp can be better than that trained by SGDM.}\label{fig:resnet18-imagenet}
\end{figure}

In this part, we discuss our experimental results on the 1000-way ImageNet classification task \cite{russakovsky2015imagenet}. We train ResNet18,
whose implementation is available at \cite{imagenet-radam}, using both SGD with momentum (SGDM, $\beta=0.9$) and ASHB with five different random seeds.
Following the common practice, we train each model for 90
epochs and decrease the learning rate by a factor of 10 at certain epochs. For SGDM, we begin with an initial learning rate of 0.1 and decay it by a factor of 10 at the 31-st and 61-st epoch, respectively.
For ASHB, we use different initial learning rates include 0.1, 0.3, 0.5, and 0.7, and we decay the learning rate at the 51-st and 71-st epochs by a factor of 10, respectively\footnote{We decay the learning rate for ASHB later than that of SGDM since ASHB plateaus slower for a given learning rate.}. Moreover, we set the weight decay to be $0.0001$ for both SGDM and ASHB.
Table~\ref{tab:acc:imagenet} lists  top-1 and top-5 accuracies of the models trained by two different optimization algorithms; we see that ASHB can outperform SGDM in classifying images. Figure~\ref{fig:resnet18-imagenet} plots the evolution of training and test accuracies. It is clear that for a given learning rate, e.g., 0.1, ASHB converges faster than SGDM. Also, according to experiments, ASHB is more robust to large learning rates, in which SGDM will blow up, but ASHB still performs well.

\subsection{Training DNNs for Natural Language Processing
}
\begin{figure*}[!htbp]
\centering
\begin{tabular}{cccc}
\hskip-0.2cm\includegraphics[width=0.24\linewidth]{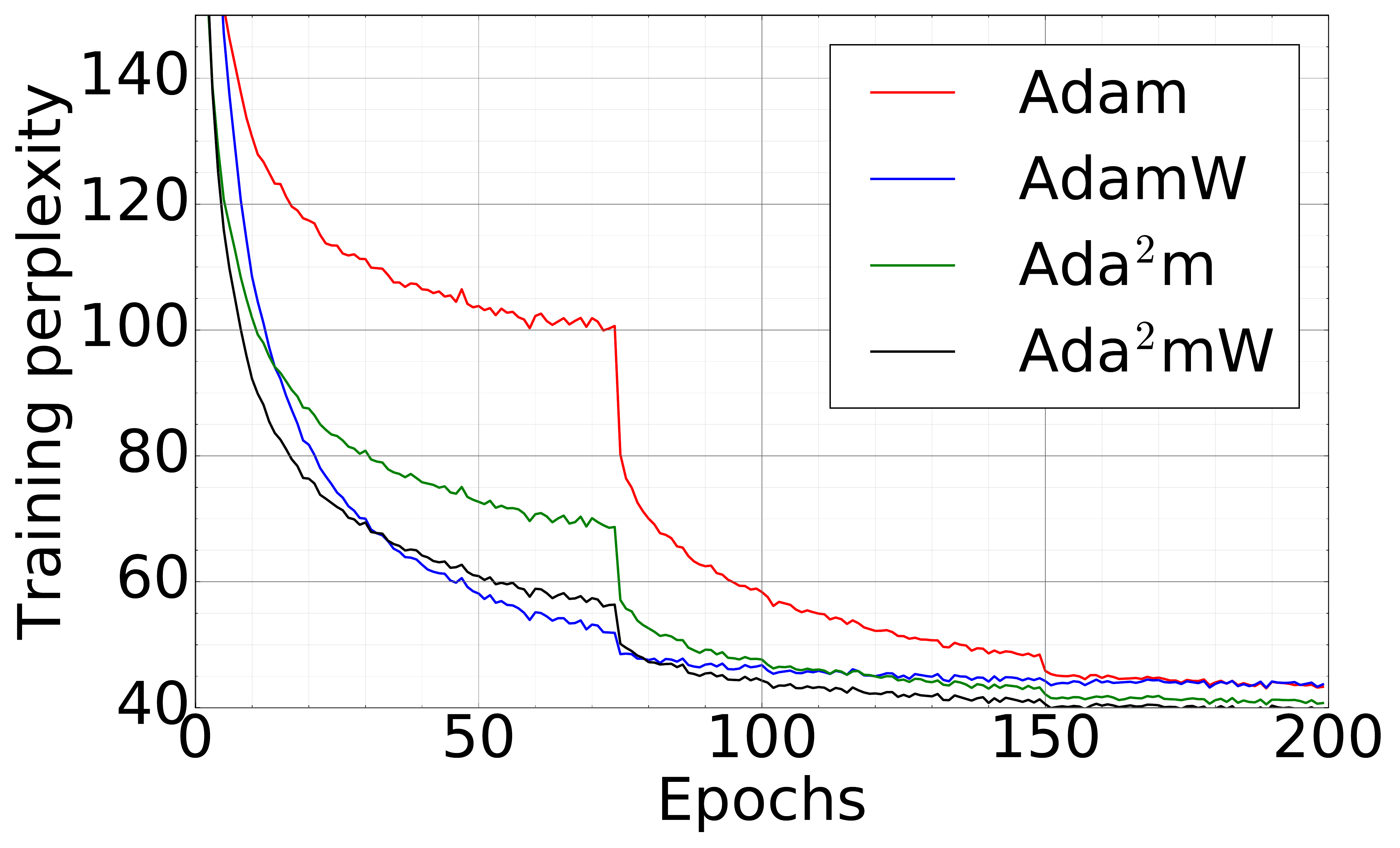}&
\hskip-0.2cm\includegraphics[width=0.24\linewidth]{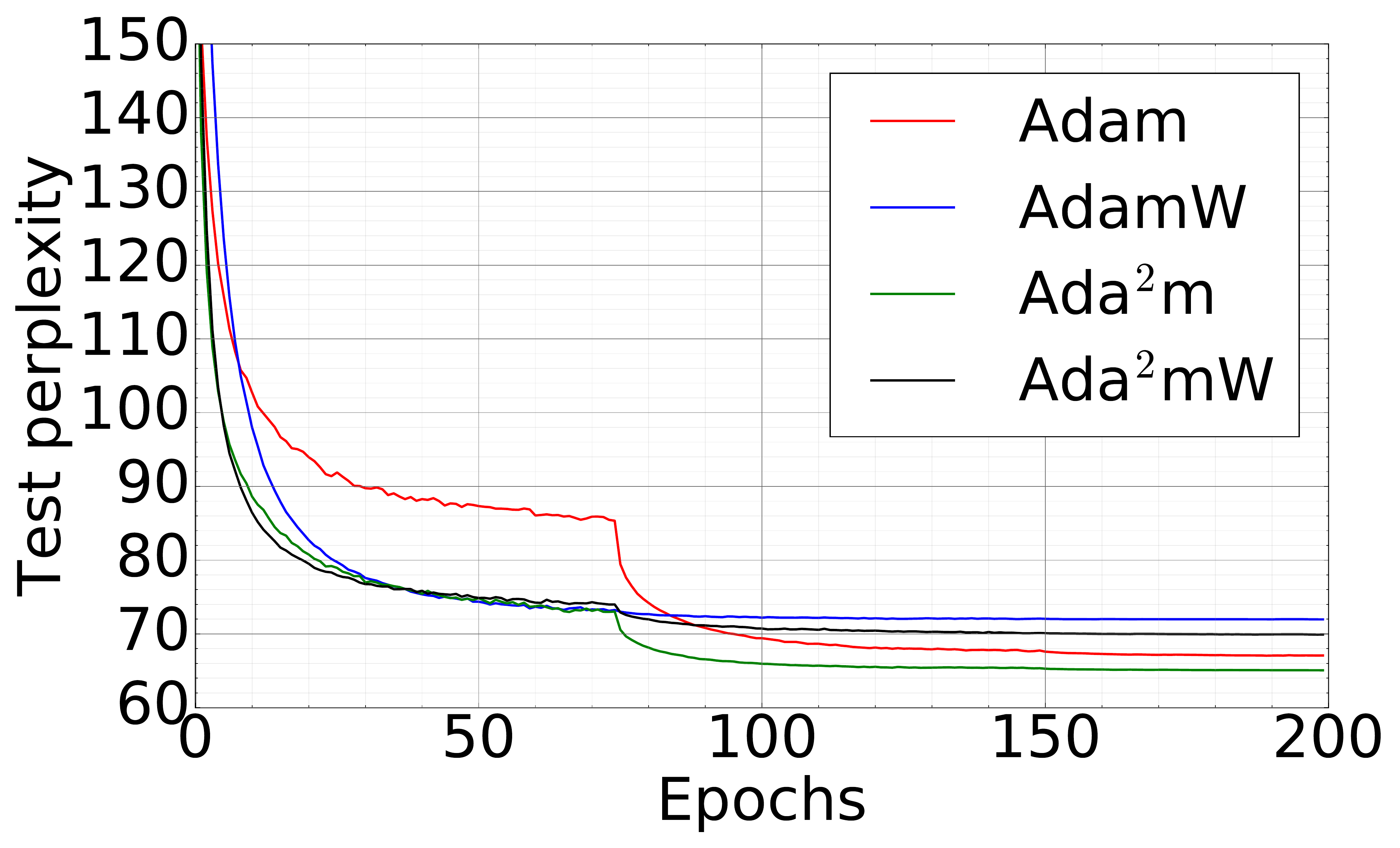}&
\hskip-0.2cm\includegraphics[width=0.24\linewidth]{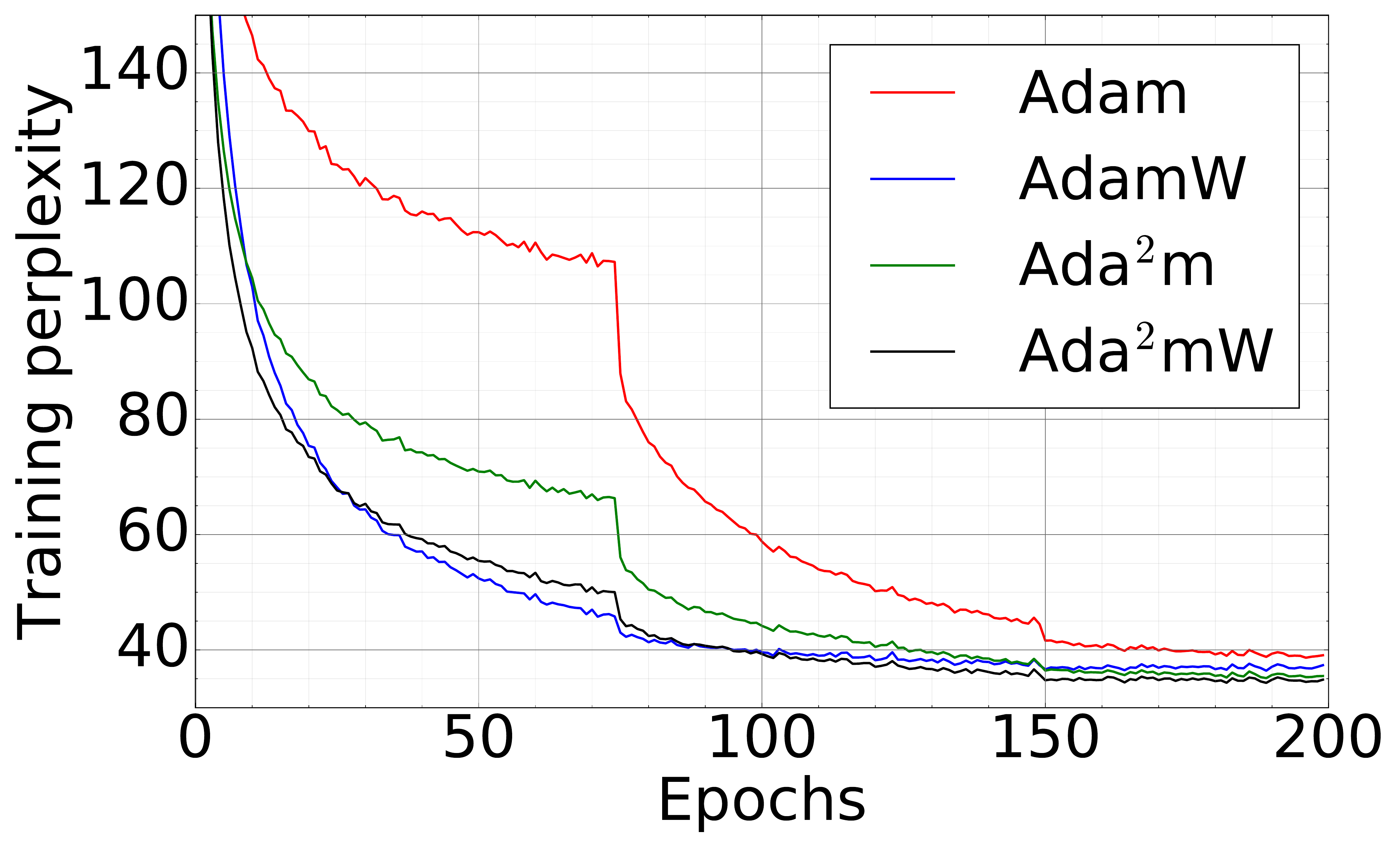}&
\hskip-0.2cm\includegraphics[width=0.24\linewidth]{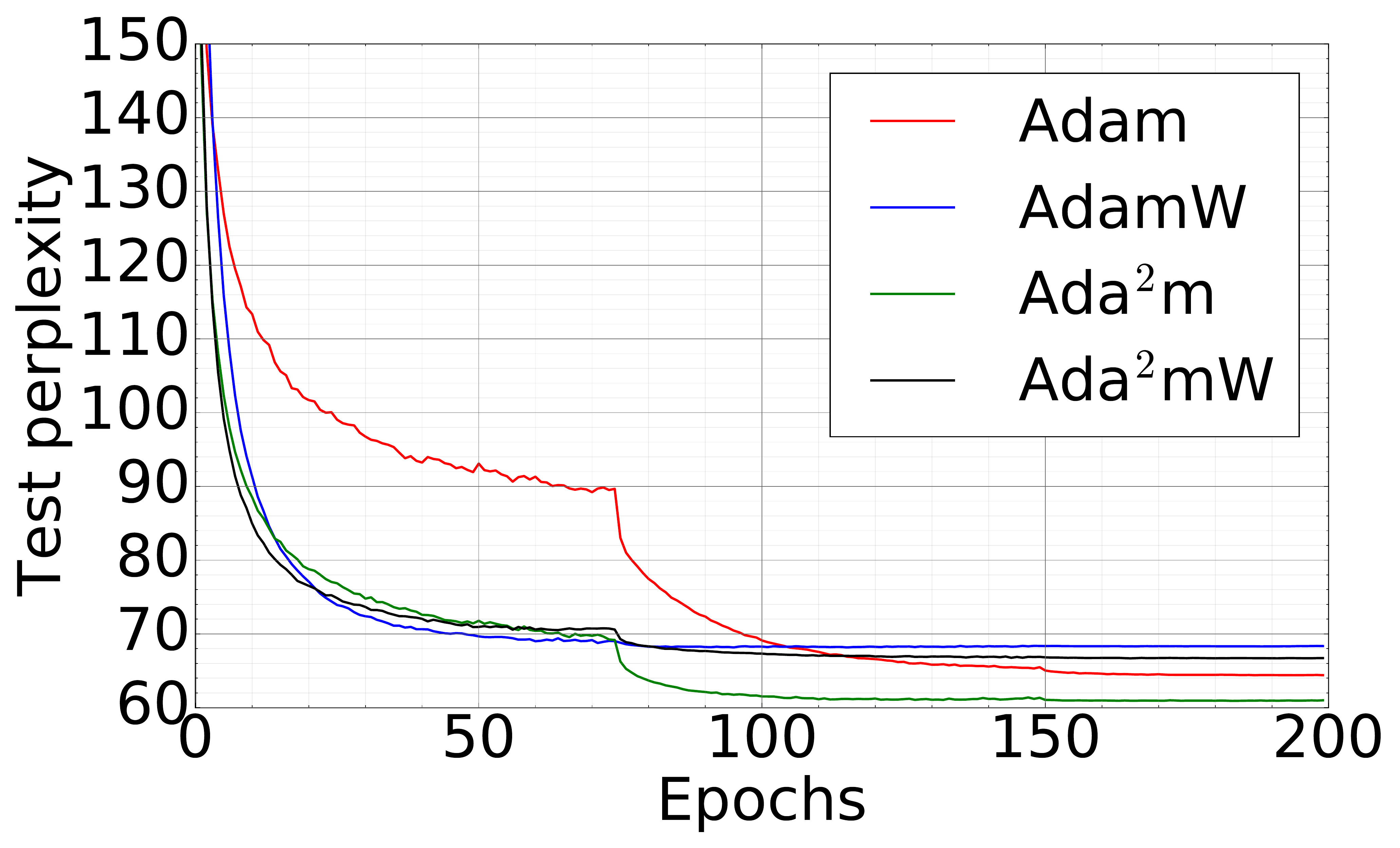}\\
\end{tabular}
\vskip -0.4cm
\caption{Training and test curves of Adam, AdamW, Ada$^2$m, and Ada$^2$mW for 2-layer (Panels 1, 2) and 3-layer LSTMs (Panels 3, 4), for the word-level Penn Treebank benchmark. Lower is better.
}
\label{fig:lstm-ptb}
\end{figure*}

\paragraph{LSTM.} We train two- and three-layer LSTM models for the benchmark word-level Penn Treebank (PTB) language modeling. We use the benchmark implementation and the default training settings of the LSTM model \cite{ptbgithub}; we switch the optimizer among Adam, AdamW, Ada$^2$m, and Ada$^2$mW. We use the fine-tuned learning rates for both Adam and AdamW, which are both $0.001$.
For Ada$^2$m and Ada$^2$mW, we set the learning rate to 0.005 and 0.003, respectively. Figure~\ref{fig:lstm-ptb} depicts the training and test curves of different optimizers. We see that the adaptive momentum schemes can not only 
accelerate training, but can also 
improve the test perplexity by a very big margin, e.g., the best test perplexities of the three-layer LSTM trained by Adam and Ada$^2$m are 64.4 and 60.9, respectively.


\paragraph{Transformers.} To further verify the efficiency of adaptive momentum for training language models, we train a transformer model for neural machine translation. Here, we train the transformer, implementation is available at \cite{transformergithub}, on the benchmark IWSLT14 De-En dataset. We note that AdamW with warmup is the default optimization algorithm for training transformers on this task; we compare AdamW and Ada$^2$mW on this task. We search the learning rate for both AdamW and Ada$^2$mW from the learning rate set $\{0.0005,0.0006,0.0007,0.0008,0.0009\}$, and we found 0.0007 is the optimal learning rate for both optimizers. We conduct experiments using five different random seeds and report their mean and standard deviation of the BLEU score (a higher BLEU score is better). For AdamW with with warmup, the mean (std) BLEU score is 35.11 (0.147) \footnote{The reported BLEU score at \cite{transformergithub} is 35.02.}. Meanwhile, the mean (std) BLEU score instead is 35.32 (0.061) for Ada$^2$mW with warmup.

\vspace{-0.4cm}
\section{Concluding Remarks}\label{sec:conclusion}
\vspace{-0.2cm}
In this paper, we propose an adaptive momentum to eliminate the computational cost for tuning the momentum-related hyperparameter for the heavy ball method. We integrate the new adaptive momentum into several benchmark algorithms, including proximal gradient descent, stochastic gradient, and Adam. Theoretically, we establish the convergence guarantee for the above benchmark algorithms with the newly developed adaptive momentum. Empirically, we see the advantage of adaptive momentum in enhancing robustness to large learning rates, accelerating training, and improving the generalization of various machine learning models for image classification and language modeling. There are numerous avenues for future work: 1) {How to interpret the adaptive momentum from the effective step size perspective \cite{10.5555/3122009.3208015,ma2018quasi}?} 2) Can we establish theoretical acceleration for the proposed adaptive momentum? 3) Can we design different optimal adaptive momentum for different optimization algorithms?



\appendices
\section{More Details of the Numerical Results}
\subsection{Detailed form of Laplacian matrix of a cyclic graph in numerical verification for Lemma \ref{le1} }\label{appendix-laplacian}
$$
{\bf L} = \begin{pmatrix}
2&-1&0&\cdots&0&-1\\
-1&2&-1&\cdots&0&0\\
0&-1&2&\cdots&0&0\\
\vdots&\vdots&\vdots&\cdots&\vdots&\vdots\\
-1&0&0&\cdots&-1&2
\end{pmatrix} \in \mathbb{R}^{d\times d},
$$

\subsection{Ablation Study--The Effects of $\delta$}\label{sec:Ablation}
\begin{figure}[!htbp]
\centering
\begin{tabular}{cc}
\hskip-0.2cm\includegraphics[width=0.5\linewidth]{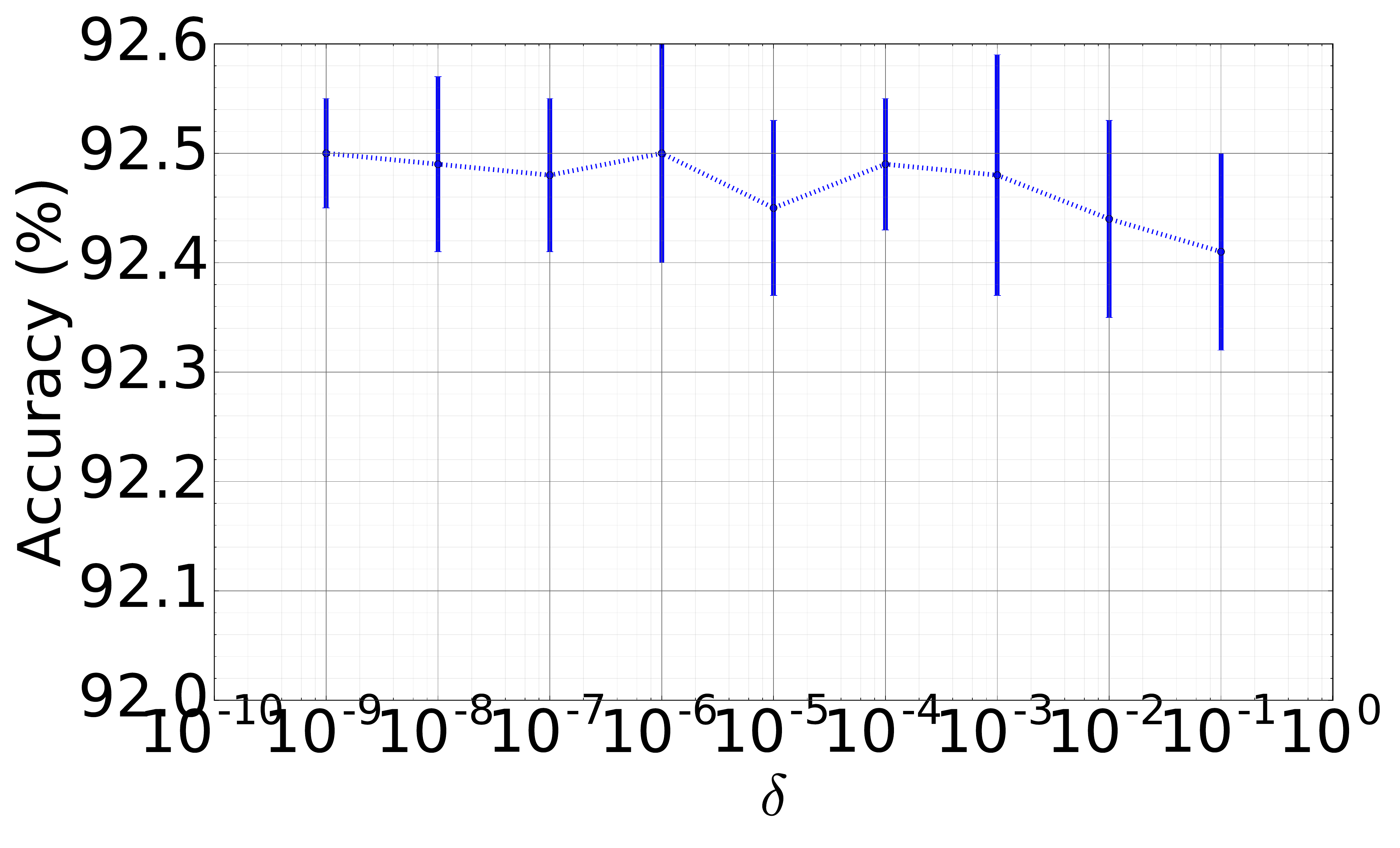}&
\hskip-0.2cm\includegraphics[width=0.5\linewidth]{figs/HBAdp_delta_v2.pdf}\\[-5pt]
ASHB & Ada$^2$mW\\
\end{tabular}
\vskip -0.3cm
\caption{$\delta$ vs. test accuracy of PreResNet20, trained by ASHB and Ada$^2$mW using different $\delta$, for CIFAR10 classification.
For both ASHB and Ada$^2$mW, we see that the classification accuracies are quite consistent for different $\delta$.
}
\label{fig:delta-effects}
\end{figure}

In this section, we study the effects of $\delta$ in the adaptive momentum \eqref{eq:adp:momentum} on the performance of ASHB and Adam$^2$W. Concerning the massive computational cost, we restrict our ablative study in training PreResNet20 for CIFAR10 classification. We test the value of $\delta$ from the set $\{10^{-i}| i=1, 2,\cdots, 9\}$. 
Figure \ref{fig:delta-effects} shows $\delta$ vs. test accuracy for ASHB and Ada$^2$mW. It is clear that the effects of the value of $\delta$ on the test accuracy of the trained models are negligible. In particular, the test accuracies are almost the same for different $\delta$ provided $\delta$ is less than $0.01$.



\section{Technical Lemmas}
\begin{lemma}
\label{lemma1}
Let $f$ have Lipschitz gradient with constant $L>0$ and let $\{{\bf x}^k\}_{k\geq 0}$ be generated by \eqref{scheme}, we have
\begin{equation}\label{lemma1-t1}
\sup_{k}\|{\bf x}^k-{\bf x}^{k-1}\| \leq \frac{\gamma R}{\delta}
\end{equation}
and
\begin{equation}
\sum_{k=1}^K \|{\bf x}^k-{\bf x}^{k-1}\|^2 \leq \frac{\gamma^2KR^2}{\delta^2}.
\end{equation}

\end{lemma}

\textbf{Proof:}
It holds that
\begin{eqnarray*}
\|\gamma {\bf g}^k\|
&\geq& \mid \|{\bf x}^{k+1}-{\bf x}^k\| - \|\beta_k({\bf x}^{k}-{\bf x}^{k-1})\|\mid \\ \nonumber
&=& \mid\|{\bf x}^{k+1}-{\bf x}^k\| - (1-\delta)\|{\bf x}^{k}-{\bf x}^{k-1}\|\mid.
\end{eqnarray*}
Thus,
\begin{align}\label{le1-t1}
\|\gamma {\bf g}^k\|^2 &\geq \left(\|{\bf x}^{k+1}-{\bf x}^k\| -(1-\delta)\|{\bf x}^k-{\bf x}^{k-1}\|\right)^2\nonumber \\
&= \|{\bf x}^{k+1}-{\bf x}^k\|^2 - 2(1-\delta)\|{\bf x}^{k+1}-{\bf x}^{k}\|\nonumber \\
&\qquad\times\|{\bf x}^k-{\bf x}^{k-1}\| + (1-\delta)^2\|{\bf x}^k-{\bf x}^{k-1}\|^2 \nonumber\\
&\geq \delta\|{\bf x}^{k+1}-{\bf x}^k\|^2 -\delta(1-\delta)\|{\bf x}^{k}-{\bf x}^{k-1}\|^2.
\end{align}
By using the Mathematical Induction (MI) method, we  then  get \eqref{lemma1-t1}.
Summing \eqref{le1-t1} from $k=1$ to $K$, we get
\begin{equation}
\delta^2\sum_{k=1}^K \|{\bf x}^k-{\bf x}^{k-1}\|^2 \leq \sum_{k=1}^{K-1} \gamma^2\|{\bf g}^{k}\| \leq \gamma^2KR^2.
\end{equation}

\begin{lemma}
\label{lemma2}
Let $f$ have Lipschitz gradient with constant $L>0$ and let $\{{\bf x}^k\}_{k\geq 0}$ be generated by \eqref{scheme}, we have
\begin{equation}
\small
\sum_{k=1}^K  \beta_k\EE\langle\nabla f({\bf x}^{k}),{\bf x}^k-{\bf x}^{k-1}\rangle\leq \frac{(1-\delta)L}{\delta}\sum_{k=1}^K\EE\|{\bf x}^k-{\bf x}^{k-1}\|^2.
\end{equation}
\end{lemma}
\textbf{Proof:}
Direct computation yields
\begin{align*}
&\langle\nabla f({\bf x}^{k}),{\bf x}^k-{\bf x}^{k-1}\rangle=\langle\nabla f({\bf x}^{k-1}),{\bf x}^k-{\bf x}^{k-1}\rangle\\
&\quad+\langle\nabla f({\bf x}^{k})-\nabla f({\bf x}^{k-1}),{\bf x}^k-{\bf x}^{k-1}\rangle\\
&\leq \langle\nabla f({\bf x}^{k-1}),{\bf x}^k-{\bf x}^{k-1}\rangle+L\|{\bf x}^k-{\bf x}^{k-1}\|^2\\
&=\langle\nabla f({\bf x}^{k-1}),-\gamma  {\bf g}^{k-1}+\beta_{k-1}({\bf x}^{k-1}-{\bf x}^{k-2})\rangle\\
&\quad+L\|{\bf x}^k-{\bf x}^{k-1}\|^2.
\end{align*}
Taking expectation, we then get
 \begin{align*}
&\EE\langle\nabla f({\bf x}^{k}),{\bf x}^k-{\bf x}^{k-1}\rangle\\
&\leq -\gamma\EE\|\nabla f({\bf x}^{k-1})\|^2+\beta_{k-1}\EE\langle\nabla f({\bf x}^{k-1}),({\bf x}^{k-1}-{\bf x}^{k-2})\rangle\\
&\quad+L\EE\|{\bf x}^k-{\bf x}^{k-1}\|^2\\
&\leq \beta_{k-1}\EE\langle\nabla f({\bf x}^{k-1}),({\bf x}^{k-1}-{\bf x}^{k-2})\rangle+L\EE\|{\bf x}^k-{\bf x}^{k-1}\|^2.
\end{align*}
With induction and the fact that ${\bf x}^{1}={\bf x}^{0}$, we then get
 \begin{align*}
&\beta_k\EE\langle\nabla f({\bf x}^{k}),{\bf x}^k-{\bf x}^{k-1}\rangle\leq L\sum_{i=1}^{k-1}(\prod_{j=i}^{k-1}\beta_{j})\EE\|{\bf x}^i-{\bf x}^{i-1}\|^2\\
&\leq L\sum_{i=1}^{k}(1-\delta)^{k+1-i}\EE\|{\bf x}^i-{\bf x}^{i-1}\|^2.
\end{align*}
Thus, we have
 \begin{align*}
&\sum_{1}^K \beta_k\EE\langle\nabla f({\bf x}^{k}),{\bf x}^k-{\bf x}^{k-1}\rangle \\
&\leq L\sum_{k=1}^K\sum_{i=1}^{k}(1-\delta)^{k+1-i}\EE\|{\bf x}^i-{\bf x}^{i-1}\|^2\\
&\leq\frac{(1-\delta)L}{\delta}\sum_{k=1}^K\EE\|{\bf x}^k-{\bf x}^{k-1}\|^2.
\end{align*}

\begin{lemma}
\label{lemma3}
Let $f$ have Lipschitz gradient with constant $L>0$ and be convex, and let $\{{\bf x}^k\}_{k\geq 0}$ be generated by \eqref{scheme}, we have
\begin{equation}
\sum_{k=1}^K  \beta_k\EE\langle {\bf x}^{k}-{\bf x}^{*}, {\bf x}^k-{\bf x}^{k-1}\rangle\leq \frac{(1-\delta)}{\delta}\sum_{k=1}^K\EE\|{\bf x}^k-{\bf x}^{k-1}\|^2.
\end{equation}
\end{lemma}
\textbf{Proof:}
We can have
\begin{align*}
    &\langle {\bf x}^{k}-{\bf x}^{*}, {\bf x}^k-{\bf x}^{k-1}\rangle\\
    &= \langle {\bf x}^{k-1}-{\bf x}^{*}, {\bf x}^k-{\bf x}^{k-1}\rangle+ \langle {\bf x}^{k}-{\bf x}^{k-1}, {\bf x}^k-{\bf x}^{k-1}\rangle\\
    &= -\gamma \langle {\bf x}^{k-1}-{\bf x}^{*},{\bf g}^{k-1}\rangle+\beta_{k-1}\langle {\bf x}^{k-1}-{\bf x}^{*}, {\bf x}^{k-1}-{\bf x}^{k-2}\rangle\\
    &\quad+ \langle {\bf x}^{k}-{\bf x}^{k-1}, {\bf x}^k-{\bf x}^{k-1}\rangle+ \|{\bf x}^{k}-{\bf x}^{k-1}\|^2.
\end{align*}
Taking expectation, we derive
\begin{align}\label{lemma3-t1}
    &\EE\langle {\bf x}^{k}-{\bf x}^{*}, {\bf x}^k-{\bf x}^{k-1}\rangle= -\gamma \EE\langle {\bf x}^{k-1}-{\bf x}^{*},\nabla f({\bf x}^{k-1})\rangle\nonumber\\
    &+\beta_{k-1}\EE\langle {\bf x}^{k-1}-{\bf x}^{*}, {\bf x}^{k-1}-{\bf x}^{k-2}\rangle+ \EE\|{\bf x}^{k}-{\bf x}^{k-1}\|^2\nonumber\\
    &+ \EE\langle {\bf x}^{k}-{\bf x}^{k-1}, {\bf x}^k-{\bf x}^{k-1}\rangle\nonumber\\
    &\leq \beta_{k-1}\EE\langle {\bf x}^{k-1}-{\bf x}^{*}, {\bf x}^{k-1}-{\bf x}^{k-2}\rangle+ \EE\|{\bf x}^{k}-{\bf x}^{k-1}\|^2\nonumber\\
    &+ \EE\langle {\bf x}^{k}-{\bf x}^{k-1}, {\bf x}^k-{\bf x}^{k-1}\rangle
\end{align}
where we used $-\gamma \EE\langle {\bf x}^{k-1}-{\bf x}^{*},\nabla f({\bf x}^{k-1})\rangle\leq -\EE(f({\bf x}^{k-1})-f({\bf x}^*))\leq 0$.
With induction and the fact that ${\bf x}^{1}={\bf x}^{0}$, we are then led to
 \begin{align*}
&\beta_k\EE\langle{\bf x}^{k}-{\bf x}^{*},{\bf x}^k-{\bf x}^{k-1}\rangle\leq \sum_{i=1}^{k-1}(\prod_{j=i}^{k-1}\beta_{j})\EE\|{\bf x}^i-{\bf x}^{i-1}\|^2\\
&\leq \sum_{i=1}^{k}(1-\delta)^{k+1-i}\EE\|{\bf x}^i-{\bf x}^{i-1}\|^2.
\end{align*}
Thus, we can get
 \begin{align*}
&\sum_{1}^K \beta_k\EE\langle {\bf x}^{k}-{\bf x}^{*},{\bf x}^k-{\bf x}^{k-1}\rangle \\
 &\leq \sum_{k=1}^K\sum_{i=1}^{k}(1-\delta)^{k+1-i}\EE\|{\bf x}^i-{\bf x}^{i-1}\|^2\\
 &\leq\frac{(1-\delta)}{\delta}\sum_{k=1}^K\EE\|{\bf x}^k-{\bf x}^{k-1}\|^2.
\end{align*}

\begin{lemma}\label{core0}
Assume  $\{{\bf x}^k\}_{k\geq 0}$ is generated by Ada$^2$m and conditions of \eqref{proadam} hold. Let
$\Xi_k:= \gamma_k^2 \EE \|{\bf m}^{k}/[({\bf v}^{k})^{\frac{1}{2}}]\|^2.$
We have
\begin{align*}
\sum_{k=1}^K\Xi_k\leq  \sum_{j=1}^{K-1}\frac{\gamma^2}{\delta^2}\EE\|({\bf g}^j)^2/(\hat{{\bf v}}^{j})\|_1,
\end{align*}
where $\hat{{\bf v}}^{j}:=\sum_{i=1}^j [{\bf g}^i]^2$.
\end{lemma}
\textbf{Proof:}
 Recalling ${\bf m}^k=\sum_{j=1}^{k-1}(\Pi_{i=j+1}^{k-1}\beta_{i})(1-\beta_j){\bf g}^j$ and ${\bf v}^{k}=\frac{\sum_{i=1}^k[{\bf g}^i]^2}{k}=\frac{\hat{{\bf v}}^{k}}{k}$,
we then have
\begin{align*}
&\gamma^2_k \|{\bf m}^{k}/[{\bf v}^{k}]^{\frac{1}{2}}\|^2= \sum_{i=1}^d\gamma_k^2 |{\bf m}^{k}_i/({\bf v}^{k}_i)^{\frac{1}{2}}|^2\\
&\leq\sum_{i=1}^d  \gamma^2_k |\sum_{j=1}^{k-1}(\Pi_{i=j+1}^{k-1}\beta_{i}){\bf g}^j_i/({\bf v}^{k})^{\frac{1}{2}}|^2\\
&=\sum_{i=1}^d  \gamma^2 |\sum_{j=1}^{k-1}(\Pi_{i=j+1}^{k-1}\beta_{i}){\bf g}^j_i/(\hat{{\bf v}}^{k})^{\frac{1}{2}}|^2,
\end{align*}
where we used $\gamma_k=\frac{\gamma}{\sqrt{k}}$.
That is further bounded by
\begin{align*}
&\sum_{i=1}^d  \gamma^2 |\sum_{j=1}^{k-1}(1-\delta)^{k-1-j}{\bf g}^j_i/(\hat{{\bf v}}^{k}_i)^{\frac{1}{2}}|^2\\
&\overset{a)}{\leq} \sum_{i=1}^d \gamma^2 (\sum_{j=1}^{k-1}(1-\delta)^{k-1-j})\cdot\sum_{j=1}^{k-1}(1-\delta)^{k-1-j}\frac{({\bf g}^j_i)^2}{\hat{{\bf v}}^{k}_i}\\
&\leq \sum_{i=1}^d  \gamma^2 \cdot\frac{1}{\delta}\cdot\sum_{j=1}^{k-1}(1-\delta)^{k-1-j}({\bf g}^j_i)^2/\hat{{\bf v}}^{k}_i\\
&=\frac{\gamma^2}{\delta}\cdot\sum_{j=1}^{k-1}(1-\delta)^{k-1-j}\|({\bf g}^j)^2/\hat{{\bf v}}^{k}\|_1\\
&\overset{b)}{\leq} \frac{\gamma^2}{\delta}\cdot\sum_{j=1}^{k-1}(1-\delta)^{k-1-j}\|({\bf g}^j)^2/\hat{{\bf v}}^{j}\|_1,
\end{align*}
where $a)$ uses the fact
$(\sum_{j=1}^{k-1}a_j b_j)^2\leq \sum_{j=1}^{k-1}a_j^2 \sum_{j=1}^{k-1} b_j^2$ with $a_j=(1-\delta)^{\frac{k-1-j}{2}}$ and $b_j=(1-\delta)^{\frac{k-1-j}{2}}{\bf g}^j_i/({\bf v}^{k}_i)^{\frac{1}{2}}$, and $b)$ uses $\hat{{\bf v}}^{j}_i\leq \hat{{\bf v}}^{k}_i$ as $j\leq k$. We can further get
\begin{align*}
&\sum_{k=1}^{K}\sum_{j=1}^{k-1}(1-\delta)^{k-1-j}\|({\bf g}^j)^2/\hat{{\bf v}}^{j}\|_1\\
&=\sum_{j=1}^{K-1}\sum_{k=j+1}^{K-1}(1-\delta)^{k-1-j}\|({\bf g}^j)^2/\hat{{\bf v}}^{j}\|_1\\
&\overset{c)}{\leq}\sum_{j=1}^{K-1}\sum_{k=j+1}^{K-1}(1-\delta)^{k-j-1}k\|({\bf g}^j)^2/\hat{{\bf v}}^{j}\|_1\\
&\leq \frac{\sum_{j=1}^{K-1}\|({\bf g}^j)^2/\hat{{\bf v}}^{j}\|_1}{\delta^2},
\end{align*}
where $c)$ depends on the fact that ${\bf v}^k=\frac{\sum_{i=1}^k [{\bf g}^k]^2}{k}$.
The result is proved by
combining the inequalities above.

\begin{lemma}\label{core1}
Assume $\{{\bf x}^k\}_{k\geq 0}$ is generated by Ada$^2$m. Let
$$\Upsilon_k:=\EE\left(-\gamma_k\langle\nabla f({\bf x}^{k}),  {\bf m}^k/({\bf v}^k)^{\frac{1}{2}}\rangle\right),$$
we then have the following result
$$\Upsilon_k\leq -\gamma(1-\beta_k)\EE\|[\nabla f({\bf x}^k)^2/(\hat{{\bf v}}^{k-1})^{\frac{1}{2}}\|_1+\beta_k\Upsilon_{k-1}+\Re_k,$$
with
$
\Re_k:=\beta_k L\Xi_{k-1}+\gamma\hat{R}(\hat{R}+D)\cdot\sum_{j=1}^d [\frac{1}{(\hat{{\bf v}}^{k-1}_j)^{\frac{1}{2}}}-\frac{1}{(\hat{{\bf v}}^{k}_j)^{\frac{1}{2}}}].
$
\end{lemma}
\textbf{Proof:}
With direct computations,
\begin{align*}
&\Upsilon_k=\EE\left(-\gamma\langle\nabla f({\bf x}^{k}),  {\bf m}^k/(\hat{{\bf v}}^k)^{\frac{1}{2}}\rangle\mid \chi^k\right)\\
&=\underbrace{\EE\left(-\gamma\langle\nabla f({\bf x}^{k}),  {\bf m}^k/(\hat{{\bf v}}^{k-1})^{\frac{1}{2}}\rangle\mid \chi^k\right)}_{\textrm{I}}\\
&+\underbrace{\EE\left(\gamma\langle\nabla f({\bf x}^{k}),  {\bf m}^k/(\hat{{\bf v}}^{k-1})^{\frac{1}{2}}-{\bf m}^k/(\hat{{\bf v}}^{k})^{\frac{1}{2}}\rangle\mid \chi^k\right)}_{\textrm{II}}.
\end{align*}
Now, we are going to bound the terms \textrm{I} and \textrm{II}. The Cauchy's inequality then gives us
\begin{align*}
 \textrm{II}&\leq \EE(\gamma\langle\nabla f({\bf x}^{k}),  {\bf m}^k/(\hat{{\bf v}}^{k-1})^{\frac{1}{2}}-{\bf m}^k/(\hat{{\bf v}}^{k})^{\frac{1}{2}}\rangle\mid \chi^k)\\
 &\leq \gamma\|\nabla f({\bf x}^{k})\|\cdot\|{\bf m}^k\|\cdot\sqrt{\sum_{j=1}^d (1/(\hat{{\bf v}}^{k-1}_j)^{\frac{1}{2}}-1/(\hat{{\bf v}}^{k}_j)^{\frac{1}{2}})^2}\\
 &\leq  \gamma\hat{R}(\hat{R}+D)\cdot\sum_{j=1}^d [\frac{1}{(\hat{{\bf v}}^{k-1}_j)^{\frac{1}{2}}}-\frac{1}{(\hat{{\bf v}}^{k}_j)^{\frac{1}{2}}}].
\end{align*}
where the last inequality  comes from the fact   $\hat{{\bf v}}^{k-1}_j\leq\hat{{\bf v}}^{k}_j$.
 The scheme of the algorithm gives us
\begin{align*}
\textrm{I}&=\underbrace{-\gamma(1-\beta_k)\|[\nabla f ({\bf x}^{k})]^2/(\hat{{\bf v}}^{k-1})^{\frac{1}{2}}\|_1}_{:=\P}\\
&-\gamma\beta_k\langle\nabla f({\bf x}^{k}),  {\bf m}^{k-1}/(\hat{{\bf v}}^{k-1})^{\frac{1}{2}}\rangle.
\end{align*}
Then, we have the following result
\begin{align*}
&\textrm{I}=\P-\underbrace{\gamma\beta_k \langle\nabla f({\bf x}^{k-1}),  {\bf m}^{k-1}/(\hat{{\bf v}}^{k-1})^{\frac{1}{2}}\rangle}_{:=\S}\\
&+\gamma\beta_k \langle\nabla f({\bf x}^{k-1})-\nabla f({\bf x}^{k}),  {\bf m}^{k-1}/(\hat{{\bf v}}^{k-1})^{\frac{1}{2}}\rangle\\
&\overset{a)}{\leq}\P-\S+\gamma\beta_k L \|{\bf x}^{k-1}-{\bf x}^{k}\|\cdot\| {\bf m}^{k-1}/(\hat{{\bf v}}^{k-1})^{\frac{1}{2}}\|\\
&\overset{b)}{\leq}\P-\S+\gamma^2\beta_k L \|{\bf m}^{k-1}/(\hat{{\bf v}}^{k-1})^{\frac{1}{2}}\|^2,
\end{align*}
where $a)$ uses the Cauchy's inequality and the Lipschitz continuity of $\nabla f$,  $b)$ depends on the scheme of Ada$^2$m. Taking total expectations, we get
\begin{align*}
\EE \textrm{I}&\leq-\gamma(1-\beta_k)\EE\|[\nabla f ({\bf x}^{k})]^2/(\hat{{\bf v}}^{k-1})^{\frac{1}{2}}\|_1\\
&+\beta_k\Upsilon_{k-1}+\beta_k L\Xi_{k-1}.
\end{align*}
Combination of  the inequalities \textrm{I} and \textrm{II} leads to the final result.


\section{Proof of Lemma \ref{th1}}
 Let ${\bf y}^k:=\left[\begin{array}{c}
   {\bf x}^k\\
   {\bf x}^{k-1}
\end{array}\right]$, the heavy ball can be rewritten as
\begin{align*}
     {\bf y}^{k+1}={\bf T}{\bf y}^k+ \left[\begin{array}{c}
   -\gamma {\bf b}\\
   {\bf 0}
\end{array}\right] ,
\end{align*}
where
\begin{align}\label{op}
{\bf T}=\left(
     \begin{array}{cc}
       (1+\beta)\mathbf{I}-\gamma {\bf A} & -\beta \mathbf{I} \\
      \mathbf{I} &  {\bf 0}\\
     \end{array}
   \right)
\end{align}
Let ${\bf x}^*$ be the minimizer of $f$, i.e., ${\bf A} {\bf x}^*+{\bf b}={\bf 0}$. Then, ${\bf y}^*:=\left[\begin{array}{c}
   {\bf x}^*\\
   {\bf x}^{*}
\end{array}\right]$ satisfies ${\bf T}{\bf y}^*={\bf y}^*+\left[\begin{array}{c}
   -\gamma {\bf b}\\
   {\bf 0}
\end{array}\right]$. Therefore, we can get
$${\bf y}^{k}-{\bf y}^{*}={\bf T}({\bf y}^{k-1}-{\bf y}^{*})={\bf T}^k({\bf y}^{0}-{\bf y}^{*}).$$
We turn to exploit the eigenvalues of ${\bf T}$, i.e., the complex number $\lambda$ satisfying
$$\textrm{det}\left(
     \begin{array}{cc}
       (\lambda-1-\beta)\mathbf{I}+\gamma {\bf A} & \beta \mathbf{I} \\
      -\mathbf{I} &  \lambda\mathbf{I}\\
     \end{array}
   \right)=0.$$
Notice that ${\bf T}$ is non-singular, all eigenvalues are nonzero. We are then led to
\begin{align*}
&\textrm{det}\left(
     \begin{array}{cc}
       (\lambda+\frac{\beta}{\lambda}-1-\beta)\mathbf{I}+\gamma {\bf A} & {\bf 0} \\
      -\mathbf{I} &  \lambda\mathbf{I}\\
     \end{array}
   \right)=0\\
   &\Longrightarrow \textrm{det}((\lambda+\frac{\beta}{\lambda})\mathbf{I}-[(1+\beta)\mathbf{I}-\gamma {\bf A}])=0.
\end{align*}
   Thus, for $\lambda^*$ being any eigenvalue of $(1+\beta)\mathbf{I}-\gamma {\bf A}$, we just need to consider
   \begin{equation}\label{fc}
       \lambda+\frac{\beta}{\lambda}=\lambda^*.
   \end{equation}

To guarantee the convergence, $\lambda$ is required to be $|\lambda|<1$.

\textbf{Convergence}:
a) If $0<\gamma\leq \frac{(1-\sqrt{\beta})^2}{L}$, $ {\bf 0}
\preceq\gamma {\bf A}
\preceq(1-\sqrt{\beta})^2$, and $1+\beta>\lambda^*\geq 2\sqrt{\beta}$, which means \eqref{fc} has real solutions.  The function $h(\lambda):= \lambda+\frac{\beta}{\lambda}$ is monotonic on $[\sqrt{\beta}, 1]$. Due to the fact that
$$h(\sqrt{\beta})=2\sqrt{\beta}\leq h(\lambda)=\lambda^*<1+\beta=h(\lambda),$$
equation \eqref{fc} has  one root over $(\sqrt{\beta}, 1)$. Similarly, it has  another  root in $(0,\sqrt{\beta})$.

b) When $\frac{(1-\sqrt{\beta})^2}{\nu}\leq\gamma\leq\frac{(1+\sqrt{\beta})^2}{L}$, $(1+\sqrt{\beta})^2\mathbf{I}\succeq\gamma {\bf A}\succeq(1-\sqrt{\beta})^2\mathbf{I}$. And then $- 2\sqrt{\beta}\leq\lambda^*\leq 2\sqrt{\beta}$. The equation has complex roots $x$ and $\bar{x}$, which obeys $x\bar{x}=|x|^2=\beta<1$. In this case, we need $\frac{(1-\sqrt{\beta})^2}{\nu}\gamma\leq\frac{(1+\sqrt{\beta})^2}{L}$, i.e., $\beta\geq \left(\frac{1-\sqrt{\frac{\nu}{L}}}{1+\sqrt{\frac{\nu}{L}}}\right)^2$.

\textbf{Optimal choice}: For any fixed $\gamma\leq\frac{1}{L}$, $(1+\beta)-\gamma L\leq \lambda^*\leq (1+\beta)-\gamma\nu$.

a) If $(1+\beta)-\gamma L\geq 2\sqrt{\beta}$, (that is $\beta\leq (1-\sqrt{rL})^2$), the larger root of \eqref{fc} is $$\frac{\lambda^*+\sqrt{(\lambda^*)^2-4\beta}}{2}\leq \lambda^*\leq (1+\beta)-\gamma\nu.$$
We want the right side is as small as possible and then set $\beta=0$ and $$\lambda_{\max}\leq 1-\gamma\nu$$.

b) If $(1+\beta)-\gamma \nu\leq 2\sqrt{\beta}$, (that is $\beta\geq (1-\sqrt{\gamma\nu})^2$),  The
equation has complex roots whose norms are both $\beta$. The optimal choice is then $\beta=(1-\sqrt{\gamma\nu})^2$ and then
$$\lambda_{\max}\leq 1-\sqrt{\gamma\nu}.$$
It is easy to see that $1-\sqrt{\gamma\nu}$ is smaller than $1-\gamma\nu$.

With the Gelfand's Theorem, given a small $\epsilon>0$, there exists $K(\mathbb{A},\epsilon)$ such that
$${\bf y}^{k}-{\bf y}^{*}\leq (1-\sqrt{\gamma\nu}+\epsilon)^k\|{\bf y}^{0}-{\bf y}^{*}\|$$
as $k\geq K(\mathbb{A},\epsilon)$.
\section{Proof of Lemma \ref{le1}}
Note the fact that
$$({\bf y}^{k+1}-{\bf y}^{k})={\bf T}({\bf y}^{k}-{\bf y}^{k-1}).$$
Recall the matrix \eqref{op}. For $\lambda^*$ being the largest eigenvalue of $(1+\beta)\mathbf{I}-\gamma {\bf A}$, the larger root is $\frac{\lambda^*+\sqrt{(\lambda^*)^2-4\beta}}{2}$. Thus, the largest eigenvalue of ${\bf T}$ is
$$t^*:=\frac{(1+\beta-\gamma \nu)+\sqrt{(1+\beta-\gamma\nu)^2-4\beta}}{2}.$$
Let ${\bf u}$ be  the vector satisfying $[(1+\beta)\mathbf{I}-\gamma {\bf A}]{\bf u}=\lambda^* {\bf u}$. And ${\bf u}$ corresponds the minimum eigenvalue of ${\bf A}$. Then, we can check that
\begin{align*}
    {\bf T}\left(
  \begin{array}{c}
t^*{\bf u}\\
{\bf u}\\
  \end{array}
\right)
=t^*\left(
  \begin{array}{c}
t^*{\bf u}\\
{\bf u}\\
  \end{array}
\right),
\end{align*}
which means ${\bf v}:=\left(
  \begin{array}{c}
t^*{\bf u}\\
{\bf u}\\
  \end{array}
\right)$  is the eigenvector with $t^*$. Due to that ${\bf A}$ has unique minimum eigenvalue, ${\bf T}$ has unique maximum  eigenvalue. The Jordan canonical form indicates that there exists ${\bf P}=[{\bf v},...]$ such that
 $${\bf P}
\left(
  \begin{array}{cc}
  t^*&\\
  &\Lambda\\
  \end{array}
\right)[{\bf P}]^{-1}=T,~~\lim_{k}(\Lambda/t^{*})^k={\bf 0}$$
With the definition ${\bf z}^k:=\left[\begin{array}{c}
   {\bf x}^k-{\bf x}^{k-1}\\
   {\bf x}^{k-1}-{\bf x}^{k-2}
\end{array}\right]$, we have
$${\bf z}^k={\bf T}^{k-3}{\bf z}^3={\bf P}\left(
  \begin{array}{cc}
  (t^*)^{k-3}&\\
  &[\Lambda]^{k-3}\\
  \end{array}
\right)[{\bf P}]^{-1}{\bf z}^3.$$
As $k$ is large,
\begin{align*}
&{\bf z}^k\approx{\bf P}\left(
  \begin{array}{cc}
  (t^*)^{k-3}&\\
  &0\\
  \end{array}
\right)[{\bf P}]^{-1}{\bf z}^3\\
&=\{[{\bf P}]^{-1}{\bf z}^3\}_1(t^*)^{k-3}{\bf v}\in \textrm{Range} ({\bf v}).
\end{align*}
That is also $\lim_{k}\frac{|\langle{\bf x}^k-{\bf x}^{k-1},{\bf u}\rangle|}{\|{\bf x}^k-{\bf x}^{k-1}\|\|{\bf u}\|}=1$.
Therefore, we are then led to
$$  \lim_k\frac{\|{\bf g}^k-{\bf g}^{k-1}\|}{\|{\bf x}^k-{\bf x}^{k-1}\|}=\lim_k\frac{\|A({\bf x}^k-{\bf x}^{k-1})\|}{\|{\bf x}^k-{\bf x}^{k-1}\|}=\nu.$$

\section{Proof of Theorem \ref{th2}}
Direct computations give us
\begin{align}\label{sc-t1}
&\EE\|{\bf x}^{k+1}-{\bf x}^*\|^2\nonumber\\
&=\EE\|{\bf x}^{k}-{\bf x}^*\|^2-2\gamma\EE\langle{\bf x}^{k}-{\bf x}^*,\nabla f({\bf x}^k)\rangle\nonumber\\
&\quad+\beta_k\EE\langle{\bf x}^k-{\bf x}^{k-1},{\bf x}^{k}-{\bf x}^*\rangle+\EE\|\beta_{k}({\bf x}^k-{\bf x}^{k-1})-\gamma {\bf g}^k\|^2\nonumber\\
&\leq \EE\|{\bf x}^{k}-{\bf x}^*\|^2-2\gamma(\EE f({\bf x}^k)-f({\bf x}^*))\nonumber\\
&\quad+\beta_k\EE\langle{\bf x}^k-{\bf x}^{k-1},{\bf x}^{k}-{\bf x}^*\rangle\nonumber\\
&\quad+2\EE\|{\bf x}^k-{\bf x}^{k-1}\|^2+2\gamma^2 \EE\|{\bf g}^k\|^2.
\end{align}
Note that \eqref{lemma3-t1} still holds, from which we can use the MI method to prove
$$\beta_k\EE\langle {\bf x}^{k}-{\bf x}^{*}, {\bf x}^k-{\bf x}^{k-1}\rangle\leq \frac{(1-\delta)R^2}{\delta^3}\gamma^2.$$
With the strongly convex property, \eqref{sc-t1} then gives us
\begin{align}\label{sc-t2}
&\EE\|{\bf x}^{k+1}-{\bf x}^*\|^2\leq (1-2\gamma\nu)\EE\|{\bf x}^{k}-{\bf x}^*\|^2 \nonumber\\
&+\frac{(1-\delta)R^2}{\delta^3}\gamma^2+\frac{4R^2}{\delta^2}\gamma^2.
\end{align}
Therefore, we get
$$\EE\|{\bf x}^{k}-{\bf x}^*\|\leq (1-2\gamma\nu)^k\EE\|{\bf x}^{0}-{\bf x}^*\|^2 +\frac{(1+3\delta)R^2}{2\delta^3\nu}\gamma.$$
By setting $C_1:=\frac{(1+3\delta)R^2}{2\delta^3\nu}$, we then proved the result.

\section{Proof of Theorem \ref{th3}}
The inequality \eqref{sc-t1} still holds.
Then, we get
\begin{align*}
&2\gamma(\EE f({\bf x}^k)-f({\bf x}^*))  \leq \EE\|{\bf x}^{k}-{\bf x}^*\|^2-\EE\|{\bf x}^{k+1}-{\bf x}^*\|^2\\
&+\beta_k\EE\langle{\bf x}^k-{\bf x}^{k-1},{\bf x}^{k}-{\bf x}^*\rangle\\
&+2\EE\|{\bf x}^k-{\bf x}^{k-1}\|^2+2\gamma^2 \EE\|{\bf g}^k\|^2.
\end{align*}
Summing from $k=1$ to $K$ and using Lemma \ref{lemma3}, we have
\begin{align*}
&2\gamma\sum_{k=1}^K(\EE f({\bf x}^k)-f({\bf x}^*))\\
&\leq \EE\|{\bf x}^{1}-{\bf x}^*\|^2+\frac{(1-\delta)}{\delta}\sum_{k=1}^K\EE\|{\bf x}^k-{\bf x}^{k-1}\|^2\\
&\quad+2\sum_{k=1}^K\EE\|{\bf x}^k-{\bf x}^{k-1}\|^2+2\gamma^2 \sum_{k=1}^K \EE\|{\bf g}^k\|^2\\
&\leq \EE\|{\bf x}^{1}-{\bf x}^*\|^2+\frac{(1+\delta)}{\delta}\frac{\gamma^2KR^2}{\delta^2}+2\gamma^2R^2K.
\end{align*}
The convexity of $f$ gives us
\begin{align*}
&\EE f(\frac{\sum_{k=1}^K{\bf x}^k}{K})-f({\bf x}^*)\\
&\qquad\leq \frac{\EE\|{\bf x}^{1}-{\bf x}^*\|^2}{2\gamma K}+\frac{(1+\delta)\gamma R^2}{2\delta^3}+\gamma R^2.
\end{align*}
Letting $C_2:=\frac{(1+\delta)R^2}{2\delta^3}+R^2$, we then get the result.
\section{Proof of Theorem \ref{th4}}
The Lipschitz property yields
\begin{align*}
    f({\bf x}^{k+1})&\leq f({\bf x}^{k})+\langle\nabla f({\bf x}^{k}),{\bf x}^{k+1}-{\bf x}^{k}\rangle+\frac{L}{2}\|{\bf x}^{k+1}-{\bf x}^{k}\|^2\\
       &= f({\bf x}^{k})-\gamma\langle\nabla f({\bf x}^{k}),{\bf g}^{k}\rangle\\
       &\quad+\beta_k\langle\nabla f({\bf x}^{k}),{\bf x}^k-{\bf x}^{k-1}\rangle+\frac{L}{2}\|{\bf x}^{k+1}-{\bf x}^{k}\|^2.
\end{align*}
Taking expectation, we then get
\begin{align*}
    \EE f({\bf x}^{k+1})&\leq \EE f({\bf x}^{k})-\frac{\gamma}{2}\EE\|\nabla f({\bf x}^{k})\|^2\\
    &+\EE(\beta_k\langle\nabla f({\bf x}^{k}),{\bf x}^k-{\bf x}^{k-1}\rangle)+\frac{L}{2}\EE\|{\bf x}^{k+1}-{\bf x}^{k}\|^2.
\end{align*}
With Lemmas \ref{lemma1} and \ref{lemma2}, we are then led to
\begin{align*}
    &\sum_{k=1}^K\frac{\gamma}{2}\EE\|\nabla f({\bf x}^{k})\|^2\leq -\frac{\delta\gamma}{1-\delta}\sum_{k=1}^{K-1}\EE\|\nabla f({\bf x}^{k})\|^2\\
    &+\frac{(1-\frac{\delta}{2})L}{\delta}\sum_{k=1}^{K+1}\EE\|{\bf x}^k-{\bf x}^{k-1}\|^2+f({\bf x}^1)-\min f.
\end{align*}
Therefore,
\begin{align*}
    \min_{1\leq k\leq K}\{\EE\|\nabla f({\bf x}^{k})\|^2\}&\leq \frac{4(1-\delta)(1-\frac{\delta}{2})L R^2}{\delta^3}\gamma\\
    &+\frac{2(1-\delta)(f({\bf x}^0)-\min f)}{\gamma K}.
\end{align*}
By denoting $C_3:=\frac{4(1-\delta)(1-\frac{\delta}{2})L R^2}{\delta^3}$ and $C_4:=2(1-\delta)$, we get the result.
\section{Proof of Proposition \ref{proprox}}
From the definition of the proximal map,  we have
\begin{align*}
&\gamma_k g({\bf x}^{k+1})+\frac{\|{\bf x}^{k+1}-[{\bf x}^k-\gamma_k\nabla f({\bf x}^k)+\beta_k({\bf x}^k-{\bf x}^{k-1})]\|^2}{2}\\
&\quad\leq\gamma_k g({\bf x}^{k})+\frac{\|{\bf x}^{k}-[{\bf x}^k-\gamma_k\nabla f({\bf x}^k)+\beta_k({\bf x}^k-{\bf x}^{k-1})]\|^2}{2}.
\end{align*}
After simplification, we are led to
\begin{align}\label{pro1-t2}
    &g({\bf x}^{k+1})-g({\bf x}^k)\nonumber\\
    &\leq \langle \frac{{\bf x}^{k+1}-{\bf x}^k}{2\gamma_k}+\nabla f({\bf x}^k)+\frac{\beta_k}{\gamma_k}({\bf x}^{k-1}-{\bf x}^k), {\bf x}^k-{\bf x}^{k+1}\rangle.
\end{align}
The Lipschitz continuity of $\nabla f$ tells us
\begin{equation}\label{pro1-t3}
    f({\bf x}^{k+1})-f({\bf x}^k)\leq \langle -\nabla f({\bf x}^k),{\bf x}^k-{\bf x}^{k+1}\rangle+\frac{L}{2}\|{\bf x}^{k+1}-{\bf x}^k\|^2.
\end{equation}
Combining (\ref{pro1-t2}) and (\ref{pro1-t3}), we get
\begin{align*}
    &F({\bf x}^{k+1})-F({\bf x}^{k})\overset{(\ref{pro1-t2})+(\ref{pro1-t3})}{\leq} \frac{\beta_k}{\gamma_k}\langle {\bf x}^{k}-{\bf x}^{k-1}, {\bf x}^{k+1}-{\bf x}^k\rangle\\
    &+(\frac{L}{2}-\frac{1}{2\gamma_k})\|{\bf x}^{k+1}-{\bf x}^k\|^2\nonumber\\
    &\overset{a)}{\leq}\frac{\beta_k}{2\gamma_k}\|{\bf x}^{k}-{\bf x}^{k-1}\|^2+(\frac{L}{2}-\frac{1}{2\gamma_k}+\frac{\beta_k}{2\gamma_k})\|{\bf x}^{k+1}-{\bf x}^k\|^2.
 \end{align*}
where $a)$ uses the Schwarz inequality $\langle {\bf x}^{k}-{\bf x}^{k-1}, {\bf x}^{k+1}-{\bf x}^k\rangle\leq \frac{1}{2}\|{\bf x}^{k}-{\bf x}^{k-1}\|^2+\frac{1}{2}\|{\bf x}^{k+1}-{\bf x}^k\|^2$. Summing from $k=1$ to $K$, we then obtain
\begin{align*}
&\sum_{k=1}^{K-1} (-\frac{L}{2}+\frac{1}{2\gamma_k}-\frac{\beta_k}{\gamma_k})\|{\bf x}^{k+1}-{\bf x}^k\|^2\\
&\leq F({\bf x}^{1})-\min F+(-\frac{L}{2}+\frac{1}{2\gamma_0}-\frac{\beta_0}{\gamma_0})\|{\bf x}^{1}-{\bf x}^0\|^2.
\end{align*}
Noticing that $\frac{1}{2\gamma_k}-\frac{\beta_k}{\gamma_k}-\frac{L}{2}=\frac{\frac{1}{2}-\beta_k}{\gamma_k}-\frac{L}{2}=\frac{1-c}{2c}L>0$, we then get the result.

\medskip

If $g$ is convex, we can use the K.K.T. condition, i.e.,
\begin{equation*}
    \frac{{\bf x}^k-{\bf x}^{k+1}}{\gamma_k}-\nabla f({\bf x}^k)+\frac{\beta_k}{\gamma_k}({\bf x}^k-{\bf x}^{k-1})\in \partial g({\bf x}^{k+1}).
\end{equation*}
With the convexity of $g$, we have
\begin{equation*}
\begin{aligned}
    &g({\bf x}^{k+1})-g({\bf x}^k)\\
    &\leq \big\langle \frac{{\bf x}^{k+1}-{\bf x}^k}{\gamma_k}+\nabla f({\bf x}^k)+\frac{\beta_k}{\gamma_k}({\bf x}^{k-1}-{\bf x}^k), {\bf x}^k-{\bf x}^{k+1}\big\rangle.
\end{aligned}
\end{equation*}
With same derivation, we need
\begin{align*}
   & F({\bf x}^{k+1})-F({\bf x}^{k})\leq \frac{\beta_k}{2\gamma_k}\|{\bf x}^{k}-{\bf x}^{k-1}\|^2\\
    &+(\frac{L}{2}-\frac{1}{\gamma_k}+\frac{\beta_k}{2\gamma_k})\|{\bf x}^{k+1}-{\bf x}^k\|^2.
 \end{align*}
 Similarly, we can get the summable result.

 \section{Proof of Proposition \ref{proadam}}
  With $\Upsilon_0=0$, Lemma \ref{core1} gives us
  \begin{align*}
  \Upsilon_k&\leq \sum_{j=1}^k (\prod_{i=j}^k(1-\beta_i) )\Big[-\gamma\EE\|[\nabla f({\bf x}^j)^2/(\hat{{\bf v}}^{j-1})^{\frac{1}{2}}\|_1\Big]\\
  &\quad+\sum_{j=1}^k (\prod_{i=j}^k(1-\beta_i) )\Re_j\\
  &\leq \sum_{j=1}^k (1-\delta)^{k-j}\Big[-\gamma\EE\|[\nabla f({\bf x}^j)^2/(\hat{{\bf v}}^{j-1})^{\frac{1}{2}}\|_1\Big]\\
  &\quad+\sum_{j=1}^k (1-\delta)^{k-j}\Re_j.
  \end{align*}
 Summation from $k=1$ to $K$,
   \begin{align}\label{thada-t1}
 &\sum_{k=1}^K \Upsilon_k\leq \sum_{k=1}^K\sum_{j=1}^k (1-\delta)^{k-j}\Big[-\gamma\EE\|[\nabla f({\bf x}^j)^2/(\hat{{\bf v}}^{j-1})^{\frac{1}{2}}\|_1\Big]\nonumber\\
  &\quad+\sum_{k=1}^K\sum_{j=1}^k (1-\delta)^{k-j}\Re_j\nonumber\\
  &\leq \sum_{j=1}^k \frac{1}{\delta}\Big[-\gamma\EE\|[\nabla f({\bf x}^j)^2/(\hat{{\bf v}}^{j-1})^{\frac{1}{2}}\|_1\Big]+\sum_{j=1}^k  \Re_j/\delta.
  \end{align}
With the Lipschitz continuity of $\nabla f$, at the point ${\bf x}^k$, we get
\begin{align*}
&f({\bf x}^{k+1})\leq f({\bf x}^{k})+\langle\nabla f({\bf x}^{k}), {\bf x}^{k+1}-{\bf x}^{k}\rangle+\frac{L}{2}\|{\bf x}^{k+1}-{\bf x}^{k}\|^2\\
&=f({\bf x}^{k})-\gamma_k\langle\nabla f({\bf x}^{k}),  {\bf m}^k/({\bf v}^k)^{1/2}\rangle+\frac{L}{2}\gamma_k^2 \EE \|{\bf m}^{k}/[({\bf v}^{k})^{\frac{1}{2}}]\|^2.
\end{align*}
Taking total condition expectation, we get
$
\EE f({\bf x}^{k+1}) \leq \EE f({\bf x}^{k})+\Upsilon_k+\frac{L}{2}\Xi_k.
$
That is also
\begin{align*}
\sum_{k=1}^{K-1} (-\Upsilon_k) \leq   f({\bf x}^{1})-\min f+\frac{L}{2}\sum_{k=1}^{K-1}\Xi_k.
\end{align*}
Combining \eqref{thada-t1}, we are then led to
\begin{align}
&\frac{1}{\delta}\sum_{k=1}^K \gamma \EE\|[\nabla f({\bf x}^k)]^2/(\hat{{\bf v}}^{k-1}\|_1\leq ( f({\bf x}^{1})-\min f)\nonumber\\
&+\frac{L}{2} \sum_{k=1}^{K-1}\Xi_k+\sum_{k=1}^K\Re_k/\delta+\gamma^2\frac{(\hat{R}+D)^2}{K(\pi)^{\frac{1}{2}}}.
\end{align}
 We can directly get
\begin{align*}
&\frac{L}{2} \sum_{k=1}^{K-1}\Xi_k+\sum_{k=1}^K\Re_k/\delta\leq  \frac{L}{2} \sum_{k=1}^{K-1}\Xi_k \\
&+\frac{1}{\delta}\sum_{k=1}^K\Big[\beta_k L\Xi_{k-1}+\gamma\hat{R}(\hat{R}+D)\cdot\sum_{j=1}^d [\frac{1}{(\hat{{\bf v}}^{k-1}_j)^{\frac{1}{2}}}-\frac{1}{(\hat{{\bf v}}^{k}_j)^{\frac{1}{2}}}]\Big]\nonumber\\
&\leq(\frac{1}{2}+\frac{1}{\delta})L\sum_{k=1}^{K-1}\Xi_k+\frac{(\hat{R}+D)^2}{K\delta\sqrt{\pi}}.
\end{align*}
 With [Lemma 6.9, \cite{li2018convergence}] and Lemma \ref{core0}, we have
$(\frac{1}{2}+\frac{1}{\delta})L\sum_{k=1}^{K-1}\Xi_k=(\frac{1}{2}+\frac{1}{\delta})L\sum_{j=1}^{K-1}\gamma^2\EE\|({\bf g}^j)^2/(\hat{{\bf v}}^{j})\|_1\leq (\frac{1}{2}+\frac{1}{\delta})L\gamma^2\ln\frac{(K-1)(\hat{R}+D)^2}{\pi}$.
On the other hand, notice $(\hat{{\bf v}}^{k-1})^{\frac{1}{2}}\leq \sqrt{k-1}(\hat{R}+D)$,
\begin{align*}
&\sum_{k=2}^K \gamma\EE\|[\nabla f({\bf x}^k)]^2/({\bf v}^{k-1})^{\frac{1}{2}}\|_1\geq \sum_{k=2}^K \frac{\gamma\EE\|\nabla f({\bf x}^k)\|^2}{[\sqrt{k-1}(\hat{R}+D)]}.
\end{align*}
In this case, we then derive
\begin{align*}
&\sum_{k=2}^K \frac{\gamma}{[\sqrt{k-1}(\hat{R}+D)]}\EE\|\nabla f({\bf x}^k)\|^2\leq \delta( f({\bf x}^{1})-\min f)\\
&+(\frac{\delta}{2}+1)L\gamma^2\ln\frac{(K-1)(\hat{R}+D)^2}{\pi}+\gamma^2\frac{(\hat{R}+D)^2}{K(\pi)^{\frac{1}{2}}}.
\end{align*}
With the fact that $\sum_{k=2}^K \frac{\EE\|\nabla f({\bf x}^k)\|^2}{\sqrt{k-1}}\geq 2\sqrt{K-1}\min_{1\leq k\leq K}\{\EE\|\nabla f({\bf x}^k)\|^2\}$, we then proved the result.

%

\bibliographystyle{ieeetr}

\bibliography{example_paper,ref}

\end{document}